\documentclass[letterpaper]{article} 
\usepackage{aaai2026}  
\usepackage{times}  
\usepackage{helvet}  
\usepackage{courier}  
\usepackage[hyphens]{url}  
\usepackage{graphicx} 
\usepackage{natbib}  
\usepackage{caption} 
\usepackage{enumitem} 
\usepackage{algorithm}
\usepackage{algorithmic}

\usepackage{newfloat}
\usepackage{listings}
\DeclareCaptionStyle{ruled}{labelfont=normalfont,labelsep=colon,strut=off} 
\floatstyle{ruled}
\newfloat{listing}{tb}{lst}{}
\floatname{listing}{Listing}
\usepackage{booktabs}
\usepackage{ragged2e}
\usepackage[export]{adjustbox}
\usepackage{xspace}
\usepackage{amsmath}
\usepackage{xcolor}
\newcommand{\idea}{IDeation\xspace}

\newcommand{\ourproblem}{Step-by-step Layered Design Generation\xspace}

\newcommand{\ourmethodfull}{Step-by-step Layered Design Generator\xspace}

\newcommand{\ourmethod}{{\fontfamily{cmr}\selectfont SLEDGE}\xspace}

\usepackage{cleveref}

\title{Step-by-step Layered Design Generation}
\author{
    Faizan Farooq Khan\textsuperscript{\rm 1}, K J Joseph\textsuperscript{\rm 2}, Koustava Goswami\textsuperscript{\rm 2}, \\ Mohamed Elhoseiny\textsuperscript{\rm 1}, Balaji Vasan Srinivasan\textsuperscript{\rm 2} \\
}
\affiliations{
    \textsuperscript{\rm 1}King Abdullah University of Science and Technology, 
    \textsuperscript{\rm 2}Adobe Research\\

}

\usepackage{bibentry}

\begin{document}

\newcommand{\mypartight}[1]{\noindent {\bf #1}}
\newcommand{\myparagraph}[1]{\vspace{3pt}\noindent\textbf{#1}\xspace}

\newcommand{\optional}[1]{{\color{gray}{#1}}}
\newcommand{\alert}[1]{{\color{red}{#1}}}
\newcommand{\gtt}[1]{{\color{purple}{#1}}}
\newcommand{\gray}[1]{{\color{gray}{#1}}}

\newcommand{\jos}[1]{{\color{blue}{#1}}}
\newcommand{\fai}[1]{{\color{cyan}{#1}}}
\newcommand{\moh}[1]{{\color{red}{#1}}}


\newcommand{\fig}[2][1]{\includegraphics[width=#1\linewidth]{fig/#2}}
\newcommand{\figh}[2][1]{\includegraphics[height=#1\linewidth]{fig/#2}}
\newcommand{\figa}[2][1]{\includegraphics[width=#1]{fig/#2}}
\newcommand{\figah}[2][1]{\includegraphics[height=#1]{fig/#2}}

\newcommand{\figsup}[2][1]{\includegraphics[width=#1\linewidth]{fig_supp/#2}}
\newcommand{\figsuph}[2][1]{\includegraphics[height=#1\linewidth]{fig_supp/#2}}
\newcommand{\figsupa}[2][1]{\includegraphics[width=#1]{fig_supp/#2}}
\newcommand{\figsupah}[2][1]{\includegraphics[height=#1]{fig_supp/#2}}
\newcommand{\figsupc}[2][1]{\includegraphics[width=#1\linewidth,cfbox=OliveGreen 1.5pt 1.5pt]{fig_supp/#2}}
\newcommand{\figsupw}[2][1]{\includegraphics[width=#1\linewidth,cfbox=BrickRed 1.5pt 1.5pt]{fig_supp/#2}}
\newcommand{\figsupq}[2][1]{\includegraphics[width=#1\linewidth,cfbox=OrangeFrame 1.5pt 1.5pt]{fig_supp/#2}}
\newcommand{\bfigsupc}[2][1]{\includegraphics[width=#1\linewidth,height=#1\linewidth,cfbox=OliveGreen 1.5pt 1.5pt]{fig_supp/#2}}
\newcommand{\bfigsupw}[2][1]{\includegraphics[width=#1\linewidth,height=#1\linewidth,cfbox=BrickRed 1.5pt 1.5pt]{fig_supp/#2}}
\newcommand{\bfigsupq}[2][1]{\includegraphics[width=#1\linewidth,height=#1\linewidth,cfbox=OrangeFrame 1.5pt 1.5pt]{fig_supp/#2}}

\newcommand{\figreb}[2][1]{\includegraphics[width=#1\linewidth]{fig_reb/#2}}
\newcommand{\figrebh}[2][1]{\includegraphics[height=#1\linewidth]{fig_reb/#2}}
\newcommand{\figreba}[2][1]{\includegraphics[width=#1]{fig_reb/#2}}
\newcommand{\figrebah}[2][1]{\includegraphics[height=#1]{fig_reb/#2}}
\newcommand{\figrebc}[2][1]{\includegraphics[width=#1\linewidth,cfbox=OliveGreen 1.0pt 1.0pt]{fig_reb/#2}}
\newcommand{\figrebw}[2][1]{\includegraphics[width=#1\linewidth,cfbox=BrickRed 1.0pt 1.0pt]{fig_reb/#2}}
\newcommand{\figrebq}[2][1]{\includegraphics[width=#1\linewidth,cfbox=OrangeFrame 1.0pt 1.0pt]{fig_reb/#2}}
\newcommand{\bfigrebc}[2][1]{\includegraphics[width=#1\linewidth,height=#1\linewidth,cfbox=OliveGreen 1.0pt 1.0pt]{fig_reb/#2}}
\newcommand{\bfigrebw}[2][1]{\includegraphics[width=#1\linewidth,height=#1\linewidth,cfbox=BrickRed 1.0pt 1.0pt]{fig_reb/#2}}
\newcommand{\bfigrebq}[2][1]{\includegraphics[width=#1\linewidth,height=#1\linewidth,cfbox=OrangeFrame 1.0pt 1.0pt]{fig_reb/#2}}
\newcommand{\mr}[2]{\multirow{#1}{*}{#2}}
\newcommand{\mc}[2]{\multicolumn{#1}{c}{#2}}
\newcommand{\Th}[1]{\textsc{#1}}
\newcommand{\tb}[1]{\textbf{#1}}

\newcommand{\stddev}[1]{\scriptsize{$\pm#1$}}

\newcommand{\diffup}[1]{{\color{OliveGreen}{($\uparrow$ #1)}}}
\newcommand{\diffdown}[1]{{\color{BrickRed}{($\downarrow$ #1)}}}

\newcommand{\deltaup}[1]{{\color{OliveGreen}{$\uparrow$ #1}}}
\newcommand{\deltadown}[1]{{\color{BrickRed}{$\downarrow$ #1}}}

\newcommand{\comment} [1]{{\color{orange} \Comment     #1}} 

\def\nmsp{\hspace{-6pt}}
\def\nssp{\hspace{-3pt}}
\def\nxssp{\hspace{-1pt}}
\def\zsp{\hspace{0pt}}
\def\xssp{\hspace{1pt}}
\def\ssp{\hspace{3pt}}
\def\msp{\hspace{6pt}}
\def\lsp{\hspace{12pt}}
\def\xlsp{\hspace{20pt}}

\newcommand{\head}[1]{{\smallskip\noindent\bf #1}}
\newcommand{\equ}[1]{(\ref{equ:#1})\xspace}


\newcommand{\nn}[1]{\ensuremath{\text{NN}_{#1}}\xspace}
\newcommand{\eq}[1]{(\ref{eq:#1})\xspace}


\def\l1{\ensuremath{\ell_1}\xspace}
\def\l2{\ensuremath{\ell_2}\xspace}


\newcommand{\tran}{^\top}
\newcommand{\mtran}{^{-\top}}
\newcommand{\zcol}{\mathbf{0}}
\newcommand{\zrow}{\zcol\tran}

\newcommand{\ind}{\mathds{1}}
\newcommand{\expect}{\mathbb{E}}
\newcommand{\nat}{\mathbb{N}}
\newcommand{\zahl}{\mathbb{Z}}
\newcommand{\real}{\mathbb{R}}
\newcommand{\proj}{\mathbb{P}}
\newcommand{\prob}{\mathbf{Pr}}

\newcommand{\mif}{\textrm{if }}
\newcommand{\other}{\textrm{otherwise}}
\newcommand{\minimize}{\textrm{minimize }}
\newcommand{\maximize}{\textrm{maximize }}

\newcommand{\id}{\operatorname{id}}
\newcommand{\const}{\operatorname{const}}
\newcommand{\sgn}{\operatorname{sgn}}
\newcommand{\erf}{\operatorname{erf}}
\newcommand{\var}{\operatorname{Var}}
\newcommand{\mean}{\operatorname{mean}}
\newcommand{\trace}{\operatorname{tr}}
\newcommand{\diag}{\operatorname{diag}}
\newcommand{\vect}{\operatorname{vec}}
\newcommand{\cov}{\operatorname{cov}}

\newcommand{\softmax}{\operatorname{softmax}}
\newcommand{\clip}{\operatorname{clip}}

\newcommand{\defn}{\mathrel{:=}}
\newcommand{\peq}{\mathrel{+\!=}}
\newcommand{\meq}{\mathrel{-\!=}}

\newcommand{\floor}[1]{\left\lfloor{#1}\right\rfloor}
\newcommand{\ceil}[1]{\left\lceil{#1}\right\rceil}
\newcommand{\inner}[1]{\left\langle{#1}\right\rangle}
\newcommand{\norm}[1]{\left\|{#1}\right\|}
\newcommand{\frob}[1]{\norm{#1}_F}
\newcommand{\card}[1]{\left|{#1}\right|\xspace}
\newcommand{\diff}{\mathrm{d}}
\newcommand{\der}[3][]{\frac{d^{#1}#2}{d#3^{#1}}}
\newcommand{\pder}[3][]{\frac{\partial^{#1}{#2}}{\partial{#3^{#1}}}}
\newcommand{\ipder}[3][]{\partial^{#1}{#2}/\partial{#3^{#1}}}
\newcommand{\dder}[3]{\frac{\partial^2{#1}}{\partial{#2}\partial{#3}}}

\newcommand{\wb}[1]{\overline{#1}}
\newcommand{\wt}[1]{\widetilde{#1}}

\newcommand{\cA}{\mathcal{A}}
\newcommand{\cB}{\mathcal{B}}
\newcommand{\cC}{\mathcal{C}}
\newcommand{\cD}{\mathcal{D}}
\newcommand{\cE}{\mathcal{E}}
\newcommand{\cF}{\mathcal{F}}
\newcommand{\cG}{\mathcal{G}}
\newcommand{\cH}{\mathcal{H}}
\newcommand{\cI}{\mathcal{I}}
\newcommand{\cJ}{\mathcal{J}}
\newcommand{\cK}{\mathcal{K}}
\newcommand{\cL}{\mathcal{L}}
\newcommand{\cM}{\mathcal{M}}
\newcommand{\cN}{\mathcal{N}}
\newcommand{\cO}{\mathcal{O}}
\newcommand{\cP}{\mathcal{P}}
\newcommand{\cQ}{\mathcal{Q}}
\newcommand{\cR}{\mathcal{R}}
\newcommand{\cS}{\mathcal{S}}
\newcommand{\cT}{\mathcal{T}}
\newcommand{\cU}{\mathcal{U}}
\newcommand{\cV}{\mathcal{V}}
\newcommand{\cW}{\mathcal{W}}
\newcommand{\cX}{\mathcal{X}}
\newcommand{\cY}{\mathcal{Y}}
\newcommand{\cZ}{\mathcal{Z}}

\newcommand{\vA}{\mathbf{A}}
\newcommand{\vB}{\mathbf{B}}
\newcommand{\vC}{\mathbf{C}}
\newcommand{\vD}{\mathbf{D}}
\newcommand{\vE}{\mathbf{E}}
\newcommand{\vF}{\mathbf{F}}
\newcommand{\vG}{\mathbf{G}}
\newcommand{\vH}{\mathbf{H}}
\newcommand{\vI}{\mathbf{I}}
\newcommand{\vJ}{\mathbf{J}}
\newcommand{\vK}{\mathbf{K}}
\newcommand{\vL}{\mathbf{L}}
\newcommand{\vM}{\mathbf{M}}
\newcommand{\vN}{\mathbf{N}}
\newcommand{\vO}{\mathbf{O}}
\newcommand{\vP}{\mathbf{P}}
\newcommand{\vQ}{\mathbf{Q}}
\newcommand{\vR}{\mathbf{R}}
\newcommand{\vS}{\mathbf{S}}
\newcommand{\vT}{\mathbf{T}}
\newcommand{\vU}{\mathbf{U}}
\newcommand{\vV}{\mathbf{V}}
\newcommand{\vW}{\mathbf{W}}
\newcommand{\vX}{\mathbf{X}}
\newcommand{\vY}{\mathbf{Y}}
\newcommand{\vZ}{\mathbf{Z}}

\newcommand{\va}{\mathbf{a}}
\newcommand{\vb}{\mathbf{b}}
\newcommand{\vc}{\mathbf{c}}
\newcommand{\vd}{\mathbf{d}}
\newcommand{\ve}{\mathbf{e}}
\newcommand{\vf}{\mathbf{f}}
\newcommand{\vg}{\mathbf{g}}
\newcommand{\vh}{\mathbf{h}}
\newcommand{\vi}{\mathbf{i}}
\newcommand{\vj}{\mathbf{j}}
\newcommand{\vk}{\mathbf{k}}
\newcommand{\vl}{\mathbf{l}}
\newcommand{\vm}{\mathbf{m}}
\newcommand{\vn}{\mathbf{n}}
\newcommand{\vo}{\mathbf{o}}
\newcommand{\vp}{\mathbf{p}}
\newcommand{\vq}{\mathbf{q}}
\newcommand{\vr}{\mathbf{r}}
\newcommand{\Vs}{\mathbf{s}}
\newcommand{\vt}{\mathbf{t}}
\newcommand{\vu}{\mathbf{u}}
\newcommand{\vv}{\mathbf{v}}
\newcommand{\vw}{\mathbf{w}}
\newcommand{\vx}{\mathbf{x}}
\newcommand{\vy}{\mathbf{y}}
\newcommand{\vz}{\mathbf{z}}

\newcommand{\vone}{\mathbf{1}}
\newcommand{\vzero}{\mathbf{0}}

\newcommand{\valpha}{{\boldsymbol{\alpha}}}
\newcommand{\vbeta}{{\boldsymbol{\beta}}}
\newcommand{\vgamma}{{\boldsymbol{\gamma}}}
\newcommand{\vdelta}{{\boldsymbol{\delta}}}
\newcommand{\vepsilon}{{\boldsymbol{\epsilon}}}
\newcommand{\vzeta}{{\boldsymbol{\zeta}}}
\newcommand{\veta}{{\boldsymbol{\eta}}}
\newcommand{\vtheta}{{\boldsymbol{\theta}}}
\newcommand{\viota}{{\boldsymbol{\iota}}}
\newcommand{\vkappa}{{\boldsymbol{\kappa}}}
\newcommand{\vlambda}{{\boldsymbol{\lambda}}}
\newcommand{\vmu}{{\boldsymbol{\mu}}}
\newcommand{\vnu}{{\boldsymbol{\nu}}}
\newcommand{\vxi}{{\boldsymbol{\xi}}}
\newcommand{\vomikron}{{\boldsymbol{\omikron}}}
\newcommand{\vpi}{{\boldsymbol{\pi}}}
\newcommand{\vrho}{{\boldsymbol{\rho}}}
\newcommand{\vsigma}{{\boldsymbol{\sigma}}}
\newcommand{\vtau}{{\boldsymbol{\tau}}}
\newcommand{\vupsilon}{{\boldsymbol{\upsilon}}}
\newcommand{\vphi}{{\boldsymbol{\phi}}}
\newcommand{\vchi}{{\boldsymbol{\chi}}}
\newcommand{\vpsi}{{\boldsymbol{\psi}}}
\newcommand{\vomega}{{\boldsymbol{\omega}}}

\newcommand{\rLambda}{\mathrm{\Lambda}}
\newcommand{\rSigma}{\mathrm{\Sigma}}

\makeatletter
\DeclareRobustCommand\onedot{\futurelet\@let@token\@onedot}
\def\@onedot{\ifx\@let@token.\else.\null\fi\xspace}
\def\eg{\emph{e.g}\onedot} \def\Eg{\emph{E.g}\onedot}
\def\ie{\emph{i.e}\onedot} \def\Ie{\emph{I.e}\onedot}
\def\vs{\emph{vs\onedot}}
\def\cf{\emph{cf}\onedot} \def\Cf{\emph{C.f}\onedot}
\def\etc{\emph{etc}\onedot} \def\vs{\emph{vs}\onedot}
\def\wrt{w.r.t\onedot} \def\dof{d.o.f\onedot}
\def\etal{\emph{et al}\onedot}
\makeatother

\newcommand\rurl[1]{%
  \href{https://#1}{\nolinkurl{#1}}%
}


\makeatletter
\let\@oldmaketitle\@maketitle
\renewcommand{\@maketitle}{\@oldmaketitle
\myfigure\bigskip}
\makeatother
\newcommand\myfigure{%
  \makebox[0pt]{\hspace{17.5cm}\includegraphics[width=1\linewidth]{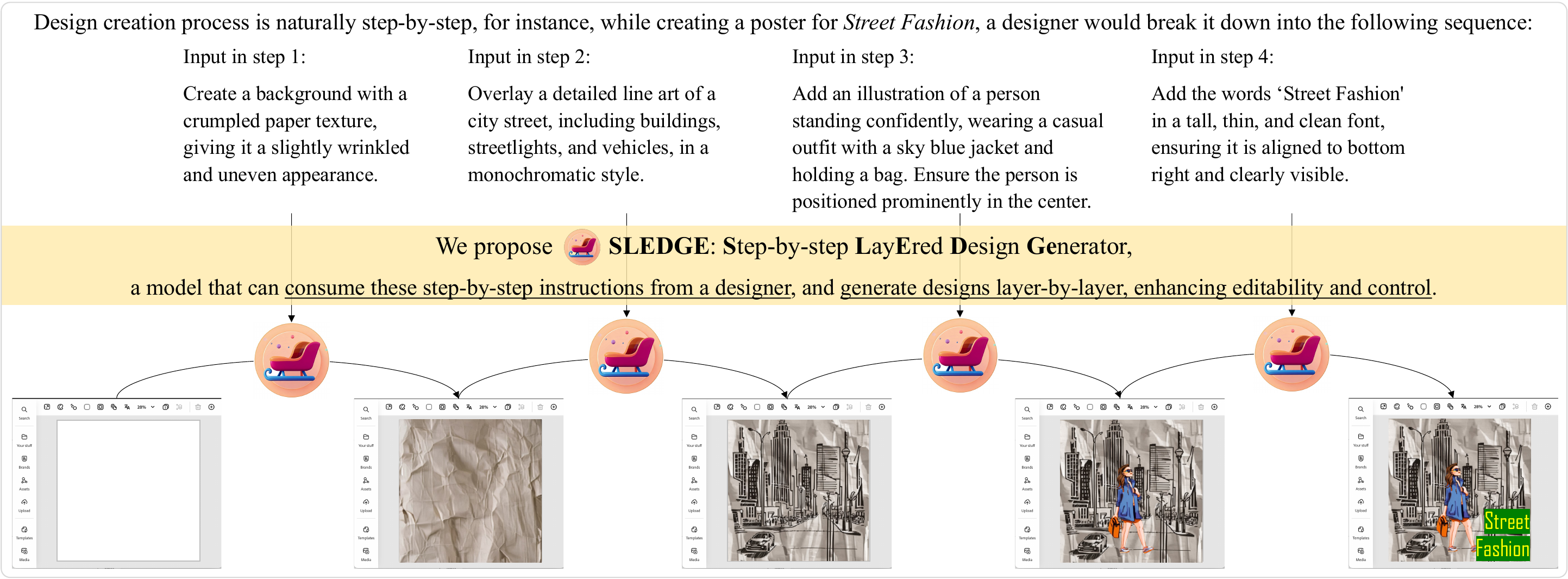}}
  \\ \refstepcounter{figure}\normalfont{Figure~\thefigure: Motivated by how design experts create graphic designs, we introduce \textit{a novel problem setup} and \textit{an approach} to generate graphic designs in a step-by-step manner, seeking inputs from the user at each step. Additionally, each step generates a layer over the previous step to allow the designer to manually intervene and adjust the design if necessary. This would facilitate Human-AI co-creation, as the AI-generated content would be naively editable using design software.
  
}
  \label{fig:teaser}
}


\maketitle

\begin{abstract}
Design generation, in its essence, is a step-by-step process where designers progressively refine and enhance their work through careful modifications. Despite this fundamental characteristic, existing approaches mainly treat design synthesis as a single-step generation problem, significantly underestimating the inherent complexity of the creative process. 
To bridge this gap, we propose a novel problem setting called Step-by-Step Layered Design Generation, which tasks a machine learning model with generating a design that adheres to a sequence of instructions from a designer. 
Leveraging recent advancements in multi-modal LLMs, we propose SLEDGE: Step-by-step LayEred Design GEnerator to model each update to a design as an atomic, layered change over its previous state, while being grounded in the instruction. To complement our new problem setting, we introduce a new evaluation suite, including a dataset and a benchmark. Our exhaustive experimental analysis and comparison with state-of-the-art approaches tailored to our new setup demonstrate the efficacy of our approach. 
We hope our work will attract attention to this pragmatic and under-explored research area.
\end{abstract}


\section{Introduction}
\label{sec:intro}
Design generation is a key application at the intersection of computer vision and creative intelligence. A design is a composition of textual and visual elements harmoniously interleaved to convey intended semantics. The typical design workflow of a designer begins with a sequence of steps (see Step 1 to Step 4 in \cref{fig:teaser}) and uses generation tools to bring the plan to life. These tools let designers manipulate element locations and modify attributes like font size. The ability to review and update is essential in the design generation cycle.

Advances in Multi-modal Large Language Models (MLLMs)~\cite{mllm1, mllm4, mllm5, mllm6, mllm7} and diffusion-based generative models~\cite{yang2024crossmodal,ruiz2022dreambooth,NEURIPS2021_49ad23d1,ddpm} have had unprecedented success in generating graphic designs \cite{cole, opencole} from textual prompts. In contrast to layout generation approaches \cite{lin2023layoutprompter,luo2024layoutllm,levi2023dlt,inoue2023towards, graphist2023hlg} which proposes only the location of elements, design generation approaches like COLE \cite{cole}, and Open-COLE \cite{opencole} proposes the content along with their layout information, making it much more closer to creating consumable designs.
Two key aspects that would further improve their applicability would be: 1) the ability to consume step-by-step instructions, similar to human workflows, and 2) the ability to create layered generations, enhancing the editability of the design.
These would facilitate the human-AI co-creation of graphic designs, where human creativity can be augmented with generative models.

Towards this end, we formalize and introduce the problem-setting of 
\textit{\ourproblem}, inspired by how humans carry out design workflows. Given a canvas state and a natural language instruction of the intended change from a designer, the design generator should be able to generate a modified canvas aligning with the instruction. To enhance the editability of designs, each modification should be atomic and layered on top of the previous state of the canvas. Textual elements should have their attribute metadata like font size, font type, and color predicted by the model, to enhance user control.

Though intuitive, iterative design generation presents a significant challenge in practice due to its dual requirements of maintaining editability and generating cohesive content. The complexity lies not only in adding new content but also in accurately preserving various design elements, shapes, and textual components of the original canvas. As demonstrated in our experiments, existing editing models~\cite{zone, ranni} often struggle to generate content beyond localized adjustments, leading to suboptimal results in scenarios requiring holistic updates. 
Also, an ideal iterative design generator should be able to consume multi-modal input (canvas state and instructions) and generate multi-modal output (modified canvas state and metadata for the modification). These characteristics make the problem setup unique, and straightforward adaptation of existing approaches fails to provide acceptable fidelity as validated in \cref{sec:res}.

To bridge this gap, we present \textit{\ourmethod: \ourmethodfull}, a novel framework by leveraging the complementary strengths of MLLMs and diffusion models to enable controlled, layerwise, iterative design generation. Our critical insight is that combining the high-level semantic understanding of MLLMs with the fine-grained generation capabilities of diffusion models enables more precise control over the design generation process. A key challenge involved in training such a composite model is curating training data. We introduce \textit{\idea}: \underline{I}terative \underline{De}sign gener\underline{ation} dataset containing triplets of current canvas state, edit instruction, and target canvas state. \idea Dataset lets the MLLM learn a unified representational space that can interoperate between the current state of the canvas $\vC_t$, the modification instruction $\vI_t$, the new state of the canvas $\vC_{t+1}$ and the associated meta-data for modification $\vM_{t+1}$. $\vM_{t+1}$ corresponding to the textual changes that can be layered on top of $\vC_t$ via deterministic text-rendering techniques, while new images added in $\vC_{t+1}$ are extracted and harmoniously integrated into $\vC_t$. See~\cref{sec:method} for details.

In order to evaluate the model on its ability to generate designs from a sequence of instructions across a diverse set of topics, we introduce \textit{\idea benchmark}, containing instructions across $1000$ varied topics. We adapt recent state-of-the-art iterative editing approaches to the problem setup and rigorously compare against them in \cref{sec:res}. We find that \ourmethod is able to maintain improved fidelity and semantic consistency when compared to them.

\noindent {To summarize, the key highlights of our work are:}
\begin{itemize}
    \item We introduce a novel problem: \textit{\ourproblem}, where graphic designs would be generated and updated based on sequential user instructions.
    \item We propose \textit{\ourmethod: \ourmethodfull}, a novel approach to address this task, leveraging multi-modal LLMs and diffusion models to achieve high-quality layered step-by-step design updates.
    \item We present a large-scale dataset 
    consisting of over $150,000$ edit instructions for training and over $20,000$ instructions for testing, specifically designed to support iterative design generation.
    \item We introduce an evaluation benchmark with more than $1,000$ unique themes and over $10,000$ detailed edit instructions to assess the design generation performance.
    \item We establish a rigorous evaluation protocol to comprehensively assess the quality of generated designs across various models, both qualitatively and quantitatively.
\end{itemize}
\section{Related Work}
\label{sec:related}
\begin{figure*}[h]
  \centering
  \includegraphics[width=0.95\textwidth]{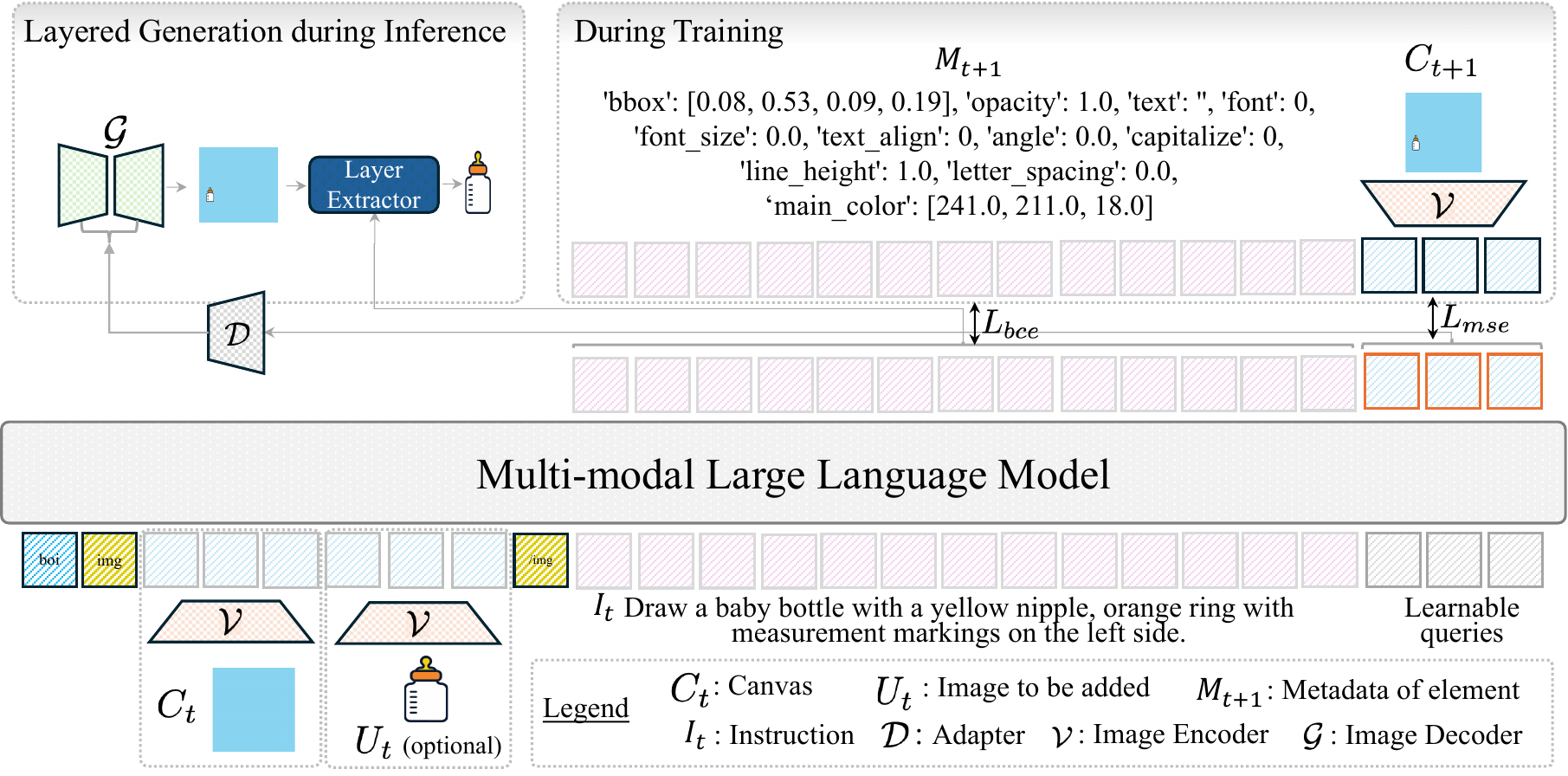}
  \caption{
  The figure provides an overview of \ourmethod: \ourmethodfull. The current state of the canvas $\vC_{t}$, the instruction from the user $\vI_{t}$, and an optional image to be inserted $\vU_{t}$ is provided to the framework. A MLLM unifies these signals to generate the next state of the $\vC_{t+1}$, along with the associated metadata, enabling layer-by-layer generation. 
  }
  \label{fig:method}
\end{figure*}

\noindent\textbf{Design Generation Approaches}: Layout generation methods~\cite{layout1, layout2, linjinpeng, layout4, crello,inoue2023layout,chen2024towards, play, layout_syneccv24, graphist2023hlg} predict element placements either conditionally or unconditionally, producing only bounding boxes with class labels (e.g., \texttt{text}, \texttt{image}, \texttt{background}). They do not generate actual content. With MLLMs, COLE~\cite{cole}, and Open-COLE~\cite{opencole} generate full designs from a caption in a single step, but cannot process sequential instructions due to their fixed input format.  In contrast, our work is among the first to enable full-spectrum design generation driven by step-by-step designer instructions.

\noindent\textbf{Iterative Generation Approaches}:
Diffusion models~\cite{ddpm} excel in generative tasks and have been adapted for iterative generation and editing~\cite{josephwacv, ranni, zone}. ZONE~\cite{zone}, extending frameworks like Instruct Pix2Pix~\cite{instructpix2pix, MagicBrush}, supports layer-by-layer edits and serves as a key baseline for our study. Integrating LLMs with diffusion models has led to significant progress~\cite{llmga, Zhou_2024_CVPR, ranni, Sun_2024_CVPR, seedx}. RANNI~\cite{ranni} uses an LLM as a planner to guide a layout-conditioned diffusion model, while LLMGA~\cite{llmga} refines prompts via LLMs and applies task-specific diffusion models for editing, inpainting, and T2I. We compare with both in \cref{sec:res}.

\noindent\textbf{Evaluation Metrics for Generated Designs}:
Evaluating generated designs is challenging. Traditional metrics focus on technical aspects like layout balance~\cite{purchase, ngo}, while recent studies emphasize perceptual and aesthetic quality. FID~\cite{fid} is widely used but has known issues: sensitivity to noise~\cite{Parmar_2022_CVPR}, diversity bias~\cite{Kynkaanniemi2022}, reliance on pre-trained models~\cite{BORJI2022103329}, and poor alignment with human perception~\cite{Chong_2020_CVPR}. To address this, we primarily adopt MLLM-based evaluation, which better captures multimodal content and aligns with human preferences~\cite{lin2024designprobe}. Haraguchi \etal~\cite{haraguchi2024can} further support this choice by showing strong correlation between MLLM evaluations and human judgment. Sec.~\ref{sub:eval} details our MLLM-based evaluation protocol. For completeness, we also report FID scores.

\section{\ourproblem}\label{sec:method}
We focus on the novel task of generating a graphic design by consuming step-by-step instructions from a designer. At each step of the design generation process, the current canvas state $\vC_t$, represented as an image, 
a user instruction $\vI_t$, and optionally an image to be inserted $\vU_{t}$ is passed to our approach \ourmethod, denoted by $\mathcal{F}(\vC_t,\vI_t, \vU_{t}; \vtheta)$ which generates the updated canvas $\vC_{t+1}$ along with its metadata $\vM_{t+1}$. As shown in \cref{fig:method}, the metadata contains bounding-box information of its location on the canvas. Text elements additionally include content and font information. 
The first canvas $\vC_0$ is initialized blank.

Design generation necessitates a rich understanding of the interplay between the images, text, their locations, and visual aspects like color and spacing. Recent advances in multi-modal learning~\cite{internlmxcomposer2_5,mllm7,mllm6,mllm4,Sun_2024_CVPR}, which learns a unified representation space for multi-modal entities, can provide the ideal implicit bias towards modeling designs. Hence, for instantiating $\mathcal{F}(\vC_t,\vI_t, \vU_{t}; \vtheta)$, we propose to use a Multi-modal LLM architecture, explained next.
\subsection{Modeling Each Canvas Generation Step} \label{sec:train}
An LLM is primarily trained to process text tokens for both input and output. However, for our specific application, we require the ability to incorporate both $\mathbf{C}_t$ (canvas representations) and $\mathbf{I}_t$ (text instructions) into the LLM’s workflow while also generating $\mathbf{C}_{t+1}$ as part of its outputs. This necessitates a unified representational space where visual elements (canvas) can be effectively integrated with textual information. The LLM must be able to interpret, manipulate, and condition on this shared space to generate coherent updates.  We propose a three-step pipeline to achieve this:

\noindent{\textbf{Step 1} \underline{Aligning Visual Encoder and Decoder}}: We encode the visual data in our pipeline using a pre-trained Vision Transformer~\cite{vit} based image encoder, denoted by $\mathcal{V}$. The encoded visual data by $\mathcal{V}$ goes via the\\

\noindent
\begin{minipage}[t]{0.11\textwidth} 
alignment module $\mathcal{D}$ and is then passed onto the pre-trained 
\end{minipage}
\begin{minipage}[t]{0.34\textwidth} 
    \centering
    \includegraphics[width=\linewidth, valign=t]{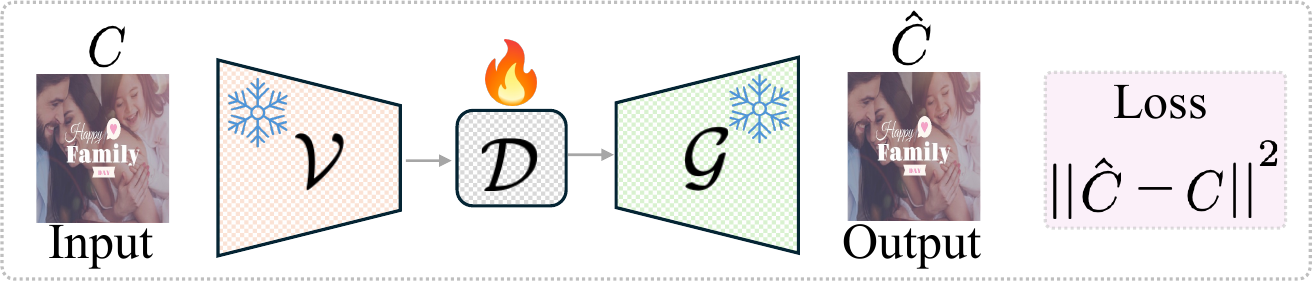}
    \captionof{figure}{Aligning encoder and decoder.} 
    \label{fig:step1}
\end{minipage}%
\hfill
  SD-XL decoder~\cite{sdxl} , denoted by $\mathcal{G}(.)$. Output of $\mathcal{D}$ replaces the textual prompts passed as input to the cross-attention layers of $\mathcal{G}$ to generate back the original canvas. The parameters of both $\mathcal{V}$ and $\mathcal{G}(.)$ are frozen, while $\mathcal{D}$ is trained with reconstruction loss, ensuring a common space for the images to be projected and generated.

\noindent{\textbf{Step 2} \underline{Aligning Visual Encoding to MLLM's Encoding}}: In this stage, we finetune the MLLM to start consuming the canvas latents from $\mathcal{V}$ along with textual data. We start off by sampling a random canvas state $\vC_t$, the text instruction $\vI_t$, and the image to be inserted $\vU_{t}$. 
Note that $\vU_{t}$ is optional. 
To simulate this during the training process, it is sampled with a probability of 0.5 and fed to $\mathcal{F}$. This mechanism enables $\mathcal{F}$ to accept additional input at inference and can act solely as a design planner. $\vC_t$ and $\vU_{t}$ are fed to $\mathcal{V}$ to get the corresponding visual embeddings before being passed onto the MLLM, as illustrated in~\cref{fig:method}.
The MLLM predicts both the metadata $\hat{\vM}_{t+1}$ and the visual embeddings of the updated canvas state $\mathcal{V}(\hat{\vC}_{t+1})$. The metadata allows for extracting the layer information as explained in~\cref{sub:inf}.

Our training approach treats metadata and visual embeddings differently. The metadata tokens, being textual in nature, are learned through next-token prediction with a cross-entropy loss. At the same time, visual embeddings use a Mean Squared Error (MSE) loss between the predicted embeddings and the embeddings of the ground truth updated canvas extracted from the ViT used to encode the initial canvas, shown in~\cref{eqn:loss}. To maintain a clear separation between visual and textual embeddings, we utilize a special $\langle$\texttt{img}$\rangle$ token that frames start and end of visual embeddings.

\begin{equation}
\label{eqn:loss}
\begin{split}
L = L_{mse}&(\mathcal{V}(\hat{\vC}_{t+1}), \mathcal{V}({\vC}_{t+1})) \\ 
    &+ L_{bce}(\hat{\vM}_{t+1}, \vM_{t+1})
\end{split}
\end{equation}

\noindent{\textbf{Step 3} \underline{Enhancing Latent Compatibility}}: In this training stage, we freeze the MLLM parameters and focus on adapting the $\mathcal{D}$ module to the MLLM's output. The function of $\mathcal{D}$ is to translate the MLLM’s output into a format interpretable by $\mathcal{G}(.)$, and is initially trained in Step 1. This learnable mod-

\noindent
\begin{minipage}[t]{0.12\textwidth}
\justifying
\noindent ule effectively enables us to utilize $\mathcal{G}(.)$ with MLLM’s multimodal o-
\end{minipage}%
\hfill
\begin{minipage}[t]{0.32\textwidth}
    \centering
    \includegraphics[width=\linewidth,valign=t]{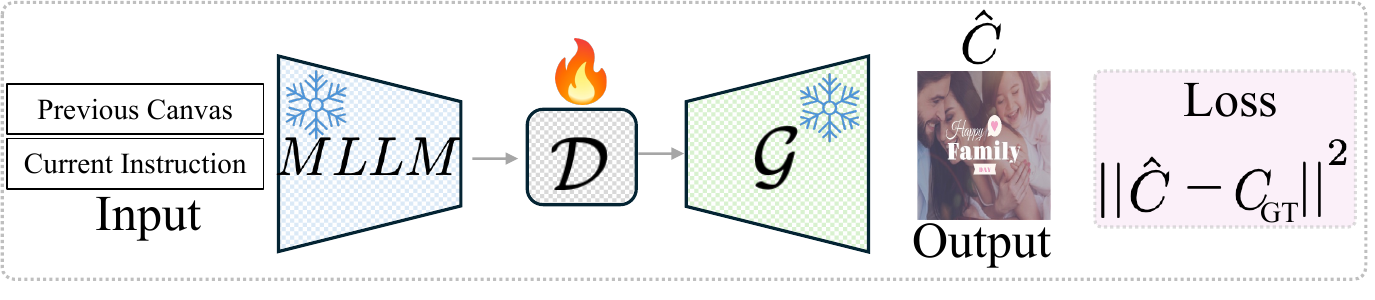}
    \captionof{figure}{Aligning MLLM and decoder.}
    \label{fig:step1_1}
\end{minipage}
utput, allowing us to process and integrate the information necessary for image synthesis. We finetune $\mathcal{D}$ by freezing the MLLM and $\mathcal{G}(.)$ parameters and propagating the reconstruction loss between $\hat{C}$ and $\mathcal{C_{GT}}$ only through the layers of $\mathcal{D}$. During training, $\mathcal{D}$ starts to understand the visual embeddings $\mathcal{V}(\hat{\vC}_{t+1})$ from the MLLM while also transforming them in a manner that can be understood by $\mathcal{G}(.)$, thus acting as a bridge between MLLM and generator.

\subsection{Facilitating Layered Generation}
\label{sub:inf}
While editing a design step-by-step, only those areas that need to be modified should be altered. The rest of the design should be consistent with the previous version of the canvas. 

\noindent\textbf{Generating the Image Layer: }
To preserve the content of $\vC_t$ while incorporating new images from $\vC_{t+1}$, we selectively extract only the edited region from $\vC_{t+1}$. For this, we leverage the bounding-box coordinates $B$ predicted by the MLLM as guidance. Using these coordinates, we generate an initial binary mask $\vM$ as defined in ~\cref{eqn:mask}, which segments the modified area from the updated canvas $\vC_{t+1}$.

\begin{equation}
    M(i, j) = 
    \begin{cases} 
        1, & \text{if } M(i, j) \text{ is inside } B, \\
        0, & \text{otherwise}.
    \end{cases}
    \label{eqn:mask}
\end{equation}

However, masks from predicted bounding boxes are often coarse and may include unintended regions. To refine them, we use Segment Anything Model (SAM)~\cite{sam} to generate high-quality masks. For each predicted mask, we compute the mean Intersection over Union(IoU) with the $\vM{=}1$ region and select the one with the highest score, filtering out spurious outputs. We then dilate the selected mask to smooth the mask boundaries. The final canvas $\vC_{t+1}$ is generated by blending the original and edited regions by:
\begin{equation}
    \vC_{t+1} = \vC_t \odot (1-\vM) + \vC_{t+1} \odot \vM,
\end{equation}
This intuitive approach ensures that only the target edit region is modified while preserving the rest of the canvas.

\noindent\textbf{Generating the Text Layer: }
Rendering text via diffusion models is challenging, often resulting in poor legibility~\cite{r22}. Given the importance of text in design, we adopt a deterministic text rendering module~\cite{opencole}, bypassing generative methods. It overlays text using font and positional cues from the MLLM, ensuring clear, layout-aligned text rendering. This ensures legible text in generated images and precise alignment with the MLLM’s predicted layout.

\section{\idea Dataset and Benchmark}
\label{sec:data}

To train and evaluate our novel problem setup, we propose \textit{\idea}: \underline{I}terative \underline{De}sign gener\underline{ation} dataset and benchmark suite. The dataset contains $182,552$ datapoints for training, complemented by $22,881$ datapoints for testing. Each data point is a triplet containing the initial state of the canvas, edit instruction, and the corresponding modified canvas. Further, in order to capture the variety of themes in which designers would create designs, we introduce a benchmark containing $10,976$ instructions across $1,066$ design themes. We provide more details below:

\subsection{\idea Dataset}
\label{sub:edit_dataset}

\begin{figure}[t]
  \centering
  \includegraphics[width=0.47\textwidth]{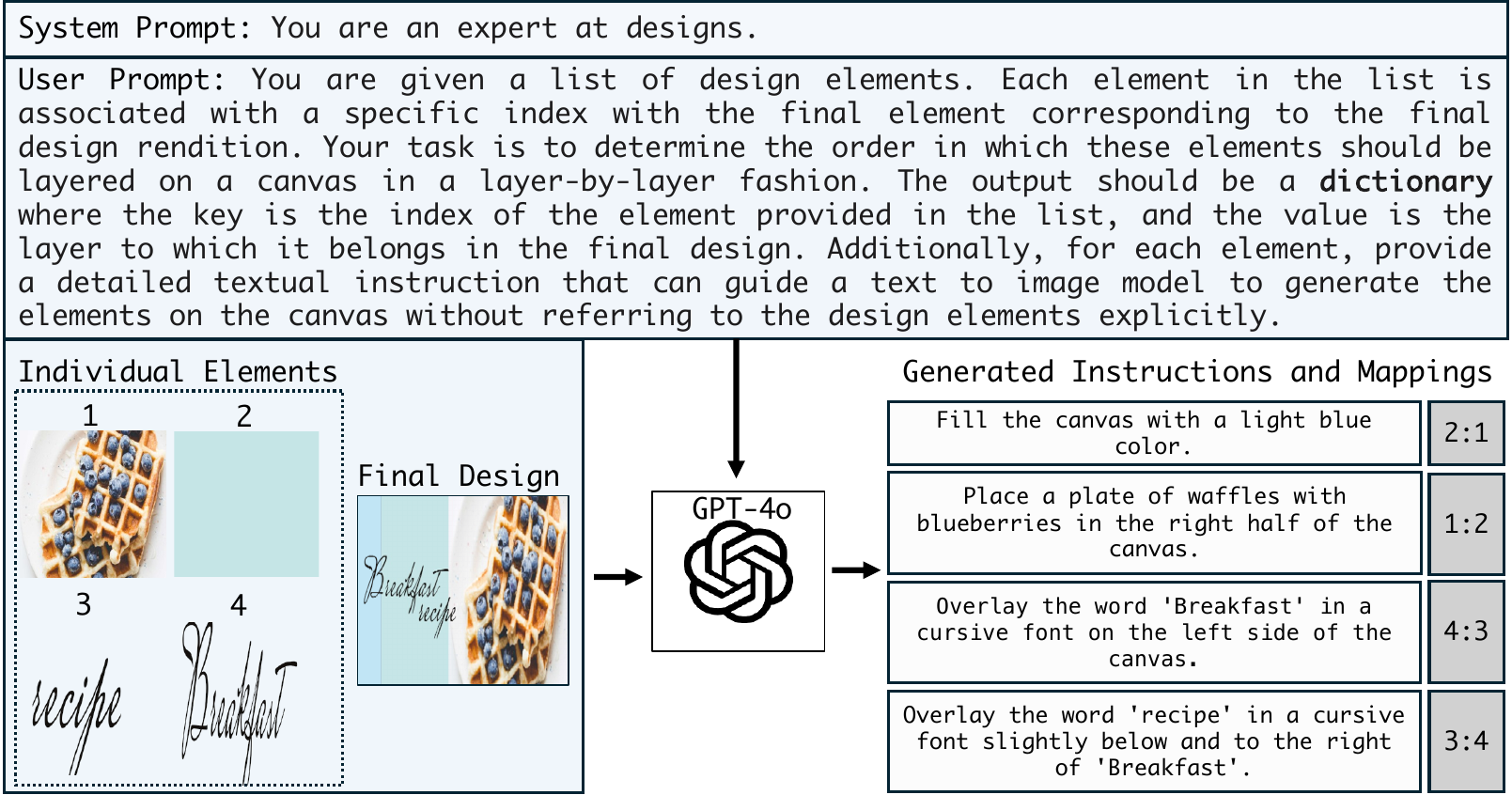}
  \caption{The figure illustrates our data generation pipeline, where each design element, along with its final composition, is processed using GPT-4o~\cite{gpt4o}. This process generates a structured mapping of layer order and detailed textual instructions that describe the transformation applied from one layer to the subsequent one.}
  \label{fig:datagen}
\end{figure}

We augment the data points in the Crello dataset~\cite{crello} with turn-by-turn edit instructions to create \idea Dataset. Crello dataset is a comprehensive resource for visual design elements sourced from the VistaCreate platform, designed to support visual synthesis tasks and layout analysis. The dataset contains over $23,000$ unique design renditions, making it one of the most extensive resources for exploring structured design composition. The dataset contains various design elements, including templates, images, icons, fonts, and layout configurations. Each design rendition is accompanied by all the individual elements associated with it. 
We extract and present to GPT-4o~\cite{gpt4o} all the individual elements and the final composite image for each design rendition in Crello. The model is then prompted to deduce the order in which elements should be placed on an empty canvas to reconstruct the design step-by-step. Alongside deciding the order of elements, GPT-4o is asked to generate textual instructions detailing each incremental update, thereby providing a dynamic reconstruction path for the final design rendition. 
Please refer to the overview of the data-generation approach in Fig.~\ref{fig:datagen}.

As a precursor to large-scale data generation, we sampled 100 design examples and manually reviewed their generated edit instructions for coherence and logical sequencing. We updated the prompts to improve the quality of the generations. Once we ensured that the generations are good for this control set, we scaled up and generated comprehensive instructions for the entire Crello~\cite{crello} dataset.

\subsection{\idea Benchmark}
While the Crello~\cite{crello} dataset offers a strong base for iterative design generation, it lacks topic diversity. Its evaluation set includes only $\sim$2,000 samples across 24 categories. To address this, we propose a broader benchmark with over 1,066 unique themes and 10,000+ detailed instructions, enabling more rigorous and diverse evaluation of model generalization across varied design scenarios.

Towards creating the \idea Benchmark, we prompt different closed-source models to generate a set of unique design themes. Specifically, we prompt Gemini~\cite{google2023gemini}, GPT-4o~\cite{gpt4o}, and Claude-Sonnet~\cite{anthropic_2024} each to give 500 different themes that can be used to create designs. We prompt different models to ensure we have a diverse set of themes. From the $1,500$ themes generated, we remove duplicates or similar themes to end-up with $1,066$ themes. They vary from ``Climate Change Awareness" to ``Tropical Beach Vacation". 

Once we have the themes, we use in-context learning and prompt GPT-4o~\cite{gpt4o} to generate edit instructions. In this way, for each theme, we have a set of instructions that we can use to generate design renditions. We evaluate the quality of edit instructions in the Ideation benchmark using GPT-4o, which confirmed $99.7\%$ of all instructions as meaningful. The remaining instructions were removed. We provide the generation prompt, in-context examples, and evaluation prompts in supplementary.

\section{Experiments and Results}
\label{sec:res}
\begin{figure*}[t]
  \centering
  \includegraphics[width=0.95\textwidth]{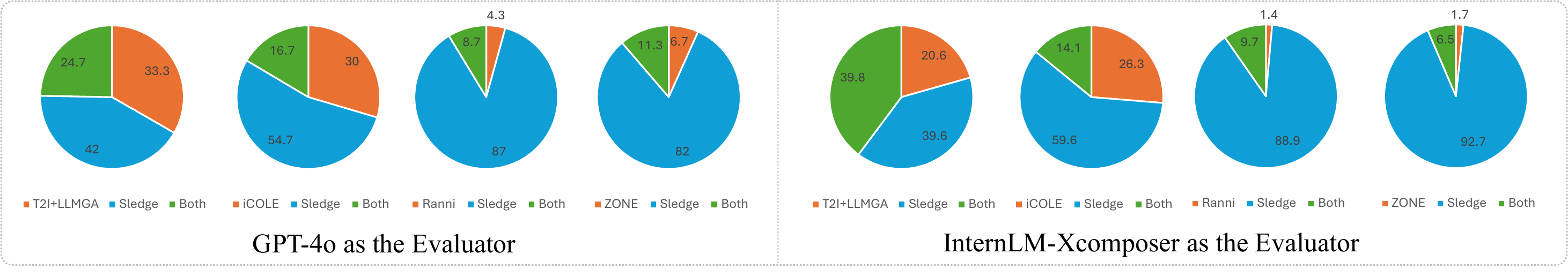}
  \caption{Performance comparison averaged across both datasets on three key aspects: theme adherence, aesthetic quality, and edit
compliance. Each baseline is compared with \ourmethod, using GPT-4o and InternLM-XComposer as the evaluators.}
  \label{fig:quan_comp}
\end{figure*}
\subsection{Baseline Methods}
As we introduce a novel problem setting, we adapt existing baselines to the new task and provide qualitative and quantitative comparisons with them. First, we adapt the state-of-the-art text-to-design approach Open-COLE \cite{opencole} to our setting in two ways, which we refer to as iCOLE and cCOLE. We also adapt sota iterative image editing methods to our setting. We explain them next:

\noindent 
\textbf{iCOLE}: We replace text-to-image module in OpenCOLE~\cite{opencole} with iterative editing module from Zone~\cite{zone}, for step-by-step generation.\\
\textbf{cCOLE}: Here, the step-by-step instructions are concatenated, and passed together into OpenCOLE~\cite{opencole}. Please note that cCOLE is not iterative like other baselines. Therefore, we compare cCOLE using metrics that measure the quality of the final rendered design in~\cref{tab:trad_metrics}.
\\\textbf{Ranni~\cite{ranni}}: combines an LLM and a diffusion model, where the LLM acts as a planner and the diffusion model as a generator. The generator is conditioned on the layout generated by the planner. It can do step-by-step editing by preserving the latents of the previous generation.
\\
\textbf{ZONE~\cite{zone}}: builds upon InstructPix2Pix~\cite{instructpix2pix}
and MagicBrush~\cite{MagicBrush} where localization information in attention maps are combined with SAM~\cite{sam} for layer-wise editing.
\\
\textbf{LLMGA~\cite{llmga}}: learns to refine input prompts, which is passed into diffusion models that can do inpainting, editing, and text-to-image generation. Using the image-editing part of LLMGA in a step-by-step manner for designs does not work well. This is because their editing model is optimized for localized edits and struggles when tasked with generating entirely new components. To alleviate this, we use their text-to-image model for the first step and then use the editing model. We refer to this as T2I+LLMGA. Please note that this is an \textit{unfair advantage} to this baseline as all other baselines and our approach \ourmethod, starts off by editing the blank canvas. 

\begin{figure*}[t]
  \centering
  \includegraphics[width=0.9\textwidth]{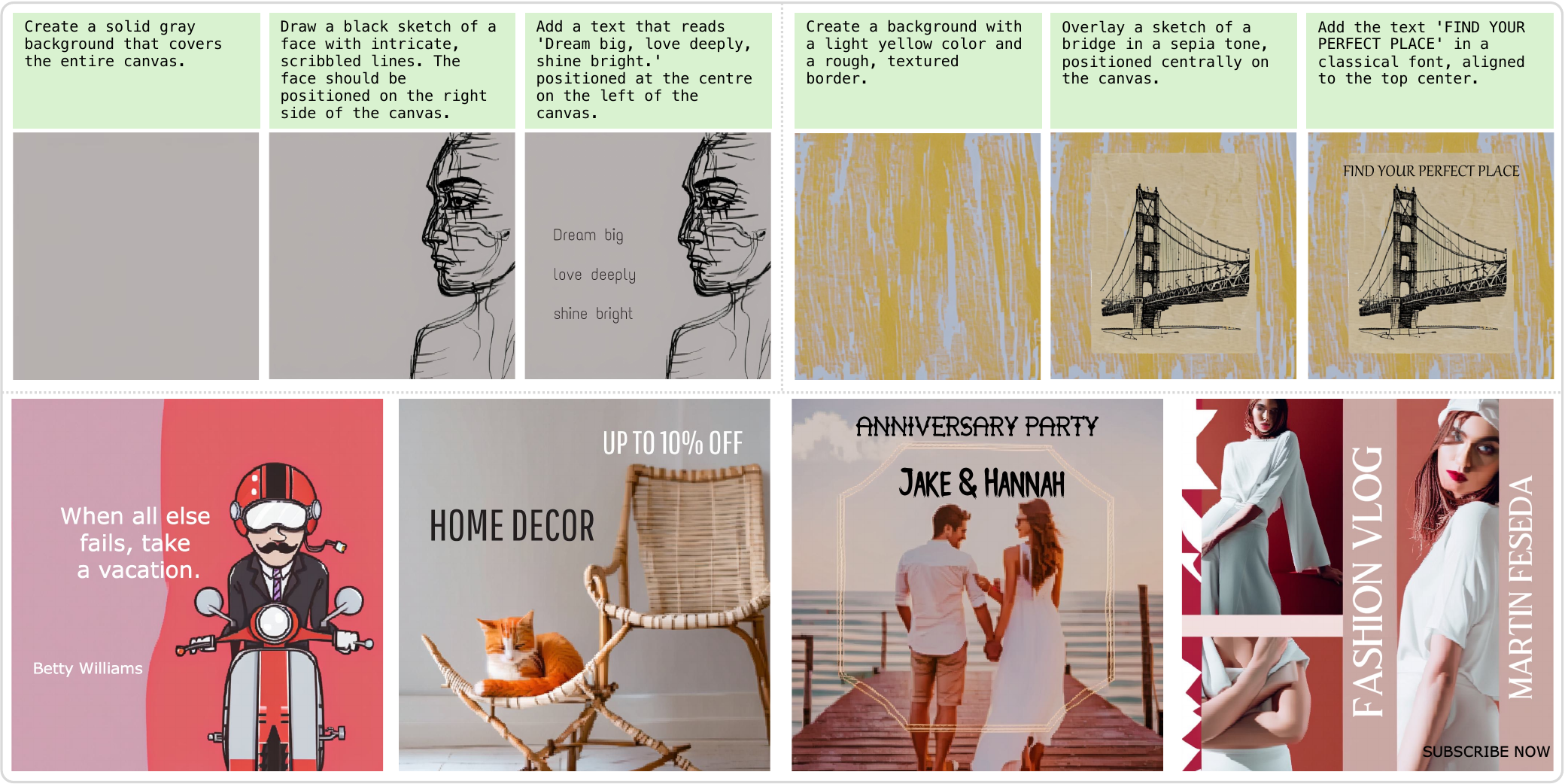}
  \caption{Illustration of iterative design generation using our method in the top row. For each design, the starting point is a blank canvas. The rest are additional final renditions from our model. Please refer to our supplementary for more examples. }
  \label{fig:design_ite}
\end{figure*}

\subsection{Evaluation Protocol}
\label{sub:eval}

We evaluate \ourmethod and baselines on three properties: 1) \textit{theme adherence}, checking alignment between the final design and the theme; 2) \textit{aesthetic quality}, evaluating the final design's aesthetics; and 3) \textit{edit compliance}, assessing how well generations follow editing instructions. These are measured either as absolute scores on a five-point Likert scale (1=poor to 5=perfect) or as relative comparisons between \ourmethod and baselines.

Following findings in~\cite{haraguchi2024can} that MLLM evaluations align well with human judgments, we adopt state-of-the-art MLLMs GPT-4o and InternLM-XComposer-2.5~\cite{internlmxcomposer2_5} as evaluators. For theme adherence, the MLLM receives the final design and associated design theme; for aesthetics, only the final design; and for edit compliance, the current canvas, updated canvas, and instructions. Prompts are in supplementary.

MLLM evaluators may exhibit ordering biases~\cite{zheng2024large}. To mitigate this, we use a circular evaluation strategy: each method pair is evaluated twice per output with flipped image order. This controls for bias, ensuring preferences are not order-dependent. If the MLLM's response remains unchanged, we treat the outputs as equal.

To enhance evaluation rigor, we use traditional metrics. \textit{FID} assesses visual quality of the final design. \textit{Aesthetic Score} is computed using a CLIP~\cite{clip} model fine-tuned on LAION-Aesthetics~\cite{aesthetic}. \textit{Text Accuracy} measures CLIP similarity between predicted and reference text. \textit{IoU} evaluates alignment between predicted and ground-truth text positions.

\begin{table}[b]
\resizebox{\columnwidth}{!}{
\begin{tabular}{lccc|cc}
\midrule
& \multicolumn{3}{c}{IDeation Benchmark} & \multicolumn{2}{c}{Crello} \\
& TA($\uparrow$) & AQ($\uparrow$) & EC($\uparrow$) & AQ($\uparrow$) & EC($\uparrow$) \\
\midrule
ZONE~\cite{zone} & 1.23($\pm$0.66) & 1.37($\pm$0.61) & 1.84($\pm$1.43) & 2.22($\pm$1.08) & 1.84($\pm$1.43) \\
Ranni~\cite{ranni} & 1.50($\pm$0.63) & 1.51($\pm$0.69) & 1.23($\pm$1.23) & 1.72($\pm$0.52) & 1.72($\pm$1.47)\\
iCOLE & 2.81($\pm$0.82) & 3.01($\pm$0.61) & 2.01($\pm$1.39) & 3.00($\pm$0.93) & 2.10($\pm$1.02) \\
T2I+LLMGA~\cite{llmga} & 3.45($\pm$0.97) & 3.13($\pm$0.57) & 2.95($\pm$1.46) & 3.12($\pm$0.44) & 2.87($\pm$1.69)\\
\ourmethod & \textbf{3.57($\pm$0.94)} & \textbf{3.61($\pm$0.41)} & \textbf{3.12($\pm$1.39)} & \textbf{3.41($\pm$0.46)} & \textbf{3.60($\pm$1.49)} \\
\midrule
\end{tabular}
\label{tab:baselines}
}
\caption{
We compare performance across both datasets on theme adherence (TA), aesthetic quality (AQ), and edit compliance (EC), reporting mean and standard deviation. Evaluator scores range from $1$ to $5$.
\label{tab:baseline}
}
\end{table}

\begin{table}[t]
\resizebox{\columnwidth}{!}{
\begin{tabular}{lcccc}
\midrule
& FID($\downarrow$) & Aesthetic Score($\uparrow$) & Text Accuracy($\uparrow$) & IoU($\uparrow$)\\
\midrule
T2I+LLMGA & 87.9 & 4.0 & - & -\\
cCOLE & 86.7 & 3.8 & 73.8 & 11.1\\
iCOLE & 171.3 & 3.2 & 74.3 & 4.5\\
\ourmethod & \textbf{46.8} & \textbf{4.2} & \textbf{89.4} & \textbf{23.5} \\
\midrule
\end{tabular}
\label{tab:metrics}
}
\caption{
Compared with baselines \ourmethod outperforms the best baselines across FID~\cite{fid}, Aesthetic Score, Text Accuracy, and IoU of textual elements.
\label{tab:trad_metrics}
}
\end{table}
\begin{figure}[h]
  \centering
  \includegraphics[width=0.48\textwidth]{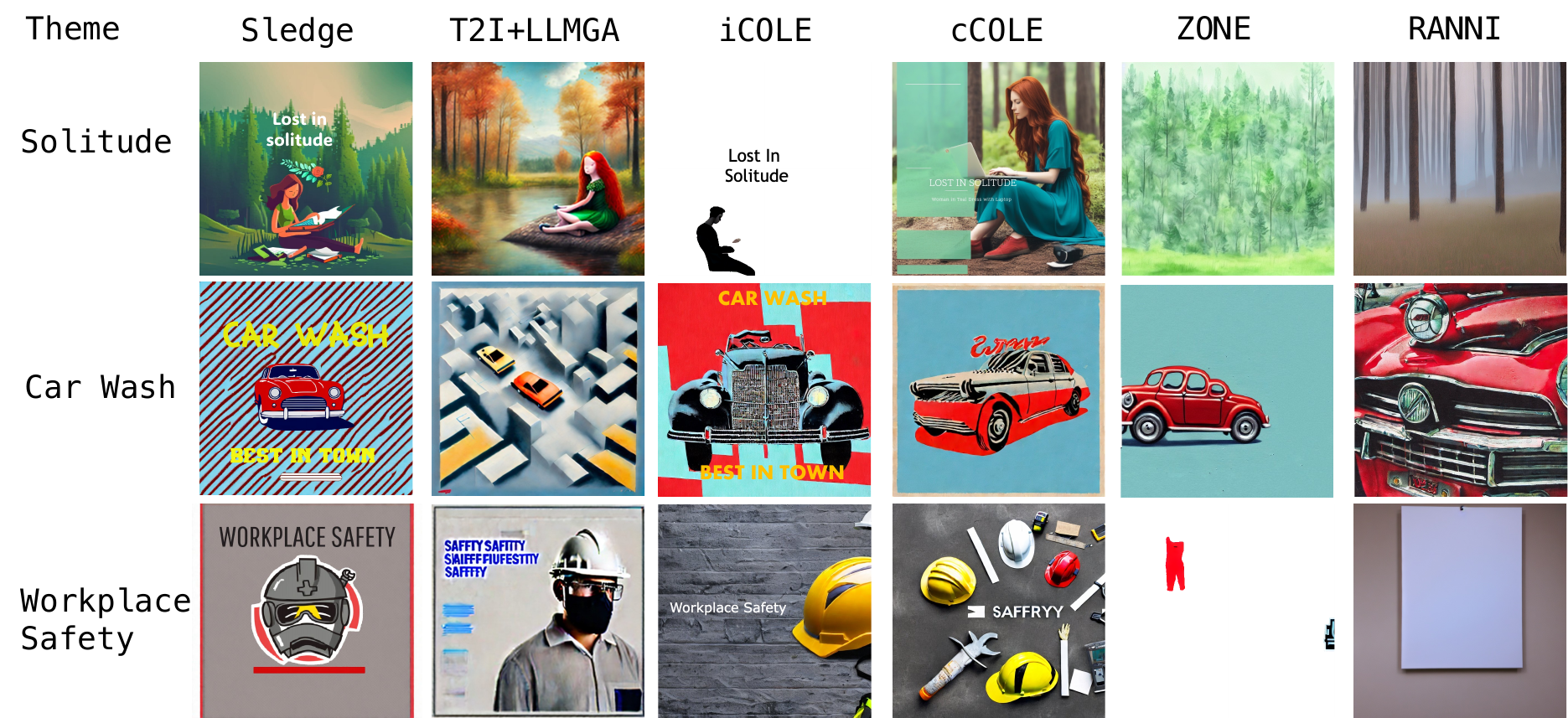}
  \caption{Qualitative comparisons between our method and baselines. The results show that Ranni and Zone and cCOLE struggle significantly with the task. While T2I-LLMGA performs better than these, it can be attributed to the better initialization from its text-to-image (T2I) model in the first step. In contrast, our approach achieves these results using a single, unified model, demonstrating more consistent performance throughout. We show more results in Supplementary.
  }
  \label{fig:qual_comp}
\end{figure}
\begin{figure}[h]
  \centering
  \includegraphics[width=0.48\textwidth]{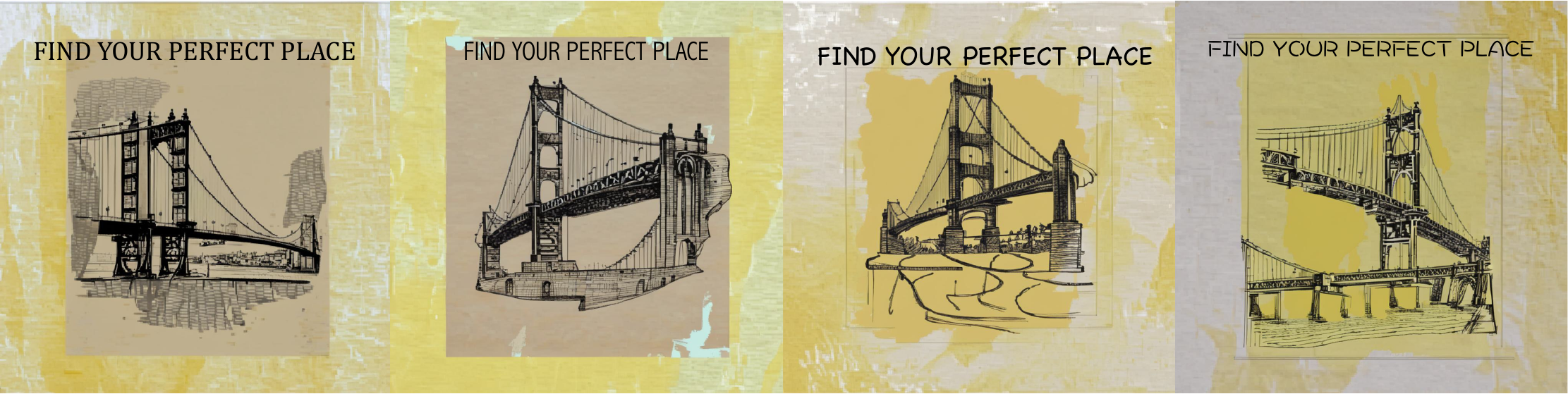}
  \caption{\ourmethod is capable of generating diverse samples for an instruction by varying the seed. The samples are generated by using the same instructions as in~\cref{fig:design_ite}.}
  \label{fig:diversity}
\end{figure}
\begin{figure}[h]
  \centering
  \includegraphics[width=0.48\textwidth]{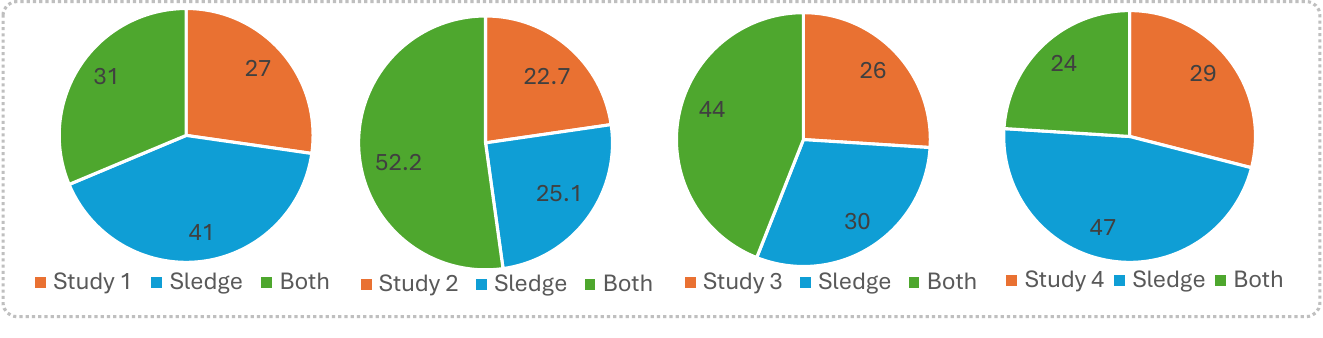}
  \caption{We compare \textit{\ourmethod} against its ablated versions, each denoted by ‘Study \#’, please refer to \cref{sec:ablation}.}
  \label{tab:ablation}
\end{figure}

\subsection{Results}

We showcase quantitative results in \cref{fig:quan_comp,tab:baseline,tab:trad_metrics} and qualitative results in \cref{fig:qual_comp,fig:design_ite,fig:diversity}.
From ~\cref{fig:qual_comp}, it can be seen that both Ranni~\cite{ranni} and ZONE~\cite{zone} fail to generate results that match the general theme of the design. Further, from~\cref{fig:quan_comp} and~\cref{tab:baseline}, it can be see that Ranni~\cite{ranni} and ZONE~\cite{zone} both perform poorly. We believe the main reason for the poor performance of ZONE~\cite{zone} could be complex edit prompts, as it is based on InstructPix2Pix~\cite{instructpix2pix} and MagicBrush~\cite{MagicBrush}, which generally excels for simpler prompts. These complex prompts can vary from localized edits to generating new objects where ZONE does not fare well. For Ranni~\cite{ranni}, the layout-conditioned model method offers a great solution for design generation. However, Ranni does not perform well on either of the evaluation datasets. As noted by others~\footnote{https://github.com/ali-vilab/Ranni/issues/7}, the model size could be a problem.

From the baselines considered, LLMGA~\cite{llmga} also suffers from similar problems as ZONE~\cite{zone}. However, with T2I+LLMGA, we can convert it into a very strong baseline. From~\cref{fig:quan_comp} and~\cref{tab:baseline}, it can be seen that T2I+LLMGA and our method show competitive performance. While cCOLE is capable of generating aesthetic outputs, it lacks the layered generation capability, making it not as useful as other approaches for design editing operations. iCOLE performs better than ZONE~\cite{zone} and RANNI~\cite{ranni}, proving to be a strong baseline, but its outputs are often not as aesthetic as \ourmethod and often are too simplistic with single objects.

Further, we perform \textbf{human evaluation} using Amazon Mechanical Turk by randomly sampling $100$ samples and comparing \ourmethod against baselines: T2I+LLMGA and iCOLE. For each sample, we evaluate theme matching, aesthetic quality, and edit compliance. 
On average, \ourmethod was preferred by 63\% of the users compared to the baselines for theme adherence. For aesthetic quality, \ourmethod was chosen by 59\% of the users, and finally, 62.5\% users preferred \ourmethod over the baselines for edit compliance.

\subsection{Ablation Studies} \label{sec:ablation}
We validate our design choices through an ablation study to assess the impact of key components in our approach. \\
\noindent{\textbf{Study 1} \underline{Effect of Fine-tuning \(\mathcal{D}\)}}: We compare our fine-tuned \(\mathcal{D}\) module against directly using the decoder after Step 1. This allows us to evaluate the benefit of adapting \(\mathcal{D}\) to our specific task. From~\cref{tab:ablation}, we see that finetuning \(\mathcal{D}\) improves the overall performance by improving alignment.\\
\noindent{\textbf{Study 2}  \underline{Full vs. Region Loss}}: We examine applying the Step 3 loss to the full predicted canvas vs. only the MLLM-specified region. The latter reduces performance, because full-image loss better leverages global context.\\
\noindent{\textbf{Study 3} \underline{\(\mathcal{U}_{t}\) handling}}: We compare removing \(\mathcal{U}_{t}\) versus replacing it with an empty canvas. Removal performs better, as empty canvas introduces misleading signals and attention overhead, while removal more clearly indicates absence.\\
\noindent{\textbf{Study 4}  \underline{Layer Extractor}}: We examine the performance of SLEDGE without the layer extractor module. This module helps to maintain consistency across the edit instructions, ensuring that the final design is according to the user's needs.  As shown in~\cref{tab:ablation}, removal of the layer extractor significantly harms the performance.

\section{Conclusion}
\label{sec:conc}
We introduce the novel problem of \textit{\ourproblem} and propose \textit{\ourmethod} to address it. To support this, we set up the \textit{\idea evaluation suite} for robust model training and evaluation. Our experiments show that existing baselines struggle with step-by-step design prompts, while \ourmethod consistently outperforms them.

However, the diversity of the design space poses unique challenges. One is handling variable resolutions, which current generative models lack but would greatly enhance creative flexibility. Another is natively supporting transparency to enable direct layer-wise generation for seamless design integration. These open directions offer promising avenues for future research, and we hope our work brings attention to this practical yet underexplored area.

\section{Acknowledgement}
This work was supported by KAUST, under Award No. BAS/1/1685-01-01.

\bibliography{aaai2026}

\appendix
\clearpage
\setcounter{page}{1}

\section*{Supplementary Material}

\noindent In this supplementary document, we include the following details that could not have been included in the main paper owing to space constraints:
\begin{itemize}[label={}]
    \item \textbf{Section \ref{sec:implem}}:  Implementation details of our approach.
    \item \textbf{Section \ref{sec:qual_res}}: Qualitative results comparing our approach with the strongest baseline.
    \item  \textbf{Section \ref{sec:prompt}:} Circular evaluation and Prompts used for dataset generation and evaluation protocol. 
    \item \textbf{Section \ref{sec:data_qual}:} Samples from our training dataset.
    \item \textbf{Section \ref{sec:fail}:} Failure cases of our model.
    \item \textbf{~\cref{fig:ctext}}: Shows a high-resolution sample of Figure 5 from the main paper focusing on the generated text.
\end{itemize}

\section{Implementation Details} \label{sec:implem}
In Step 1 of our methodology (see~\cref{sec:train}), we train the decoder module, which comprises four cross-attention layers. The output of this module replaces the original text conditioning in the pre-trained SDXL model~\cite{sdxl}. Training is conducted for 15,000 steps. The training used a per-GPU batch size of $4$, resulting in an effective batch size of $16$ across all GPUs. Optimization was performed with AdamW~\cite{adamw} using a learning rate of $10^{-4}$ and a cosine decay schedule.

In Step $2$ of the methodology, we trained the model for $12,000$ steps with a per-GPU batch size of $8$, yielding an effective batch size of $32$ ($4$ GPUs × $8$ samples each). A dropout probability of $50\%$ was applied to $\vU_{t+1}$. Our MLLM is built on top of Llama2-chat-13B~\cite{llama2} initialized with pre-trained checkpoint~\cite{seedx}. The AdamW optimizer was again employed with a $10^{-4}$ learning rate and a cosine decay schedule. The visual embeddings were extracted using the Qwen encoder $\mathcal{V}$~\cite{qwen}.

Step $3$ followed the same training protocol as step $2$, including the learning rate, optimizer, and initialization strategy, but with a reduced batch size of $4$ per GPU. This adjustment accommodated the MLLM and the diffusion model while optimizing the decoder module $\mathcal{D}$.  

All training was conducted on 4 NVIDIA A100 GPUs. 

For inference, per generation on an A100 GPU, \ourmethod takes 17.9s while iCOLE, Ranni, and T2I-LLMGA take 21.2s, 25.3s, and 17.3s, respectively.

\section{Qualitative Results} \label{sec:qual_res}
\begin{figure*}[t]
  \centering
  \includegraphics[width=0.9\textwidth]{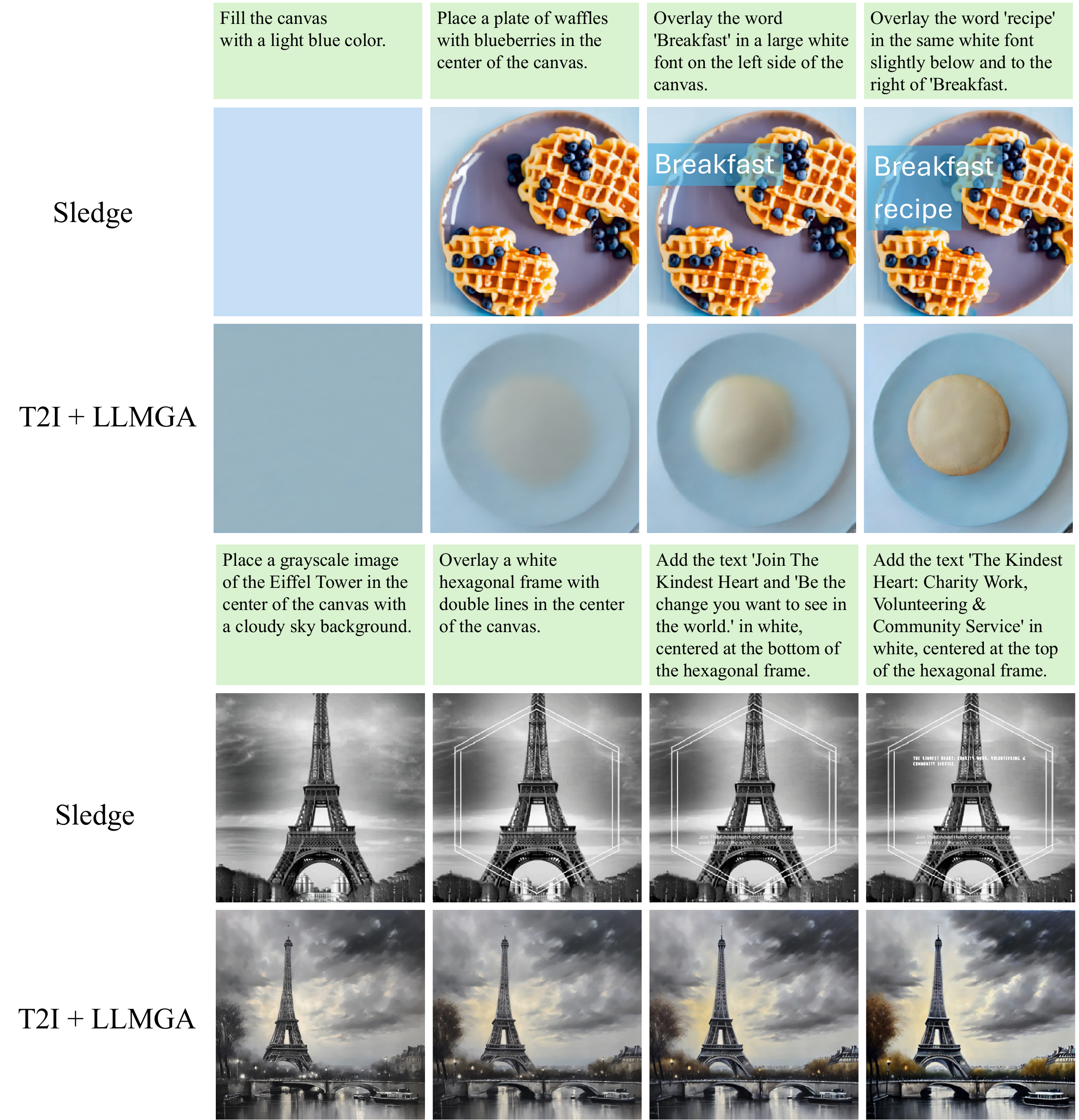}
  \caption{Qualitative comparison between our method and T2I+LLMGA for iterative design generation. This figure focuses on adding new elements to the generation process in an iterative fashion.}
  \label{fig:qual2}
\end{figure*}

\begin{figure*}[t]
  \centering
  \includegraphics[width=0.9\textwidth]{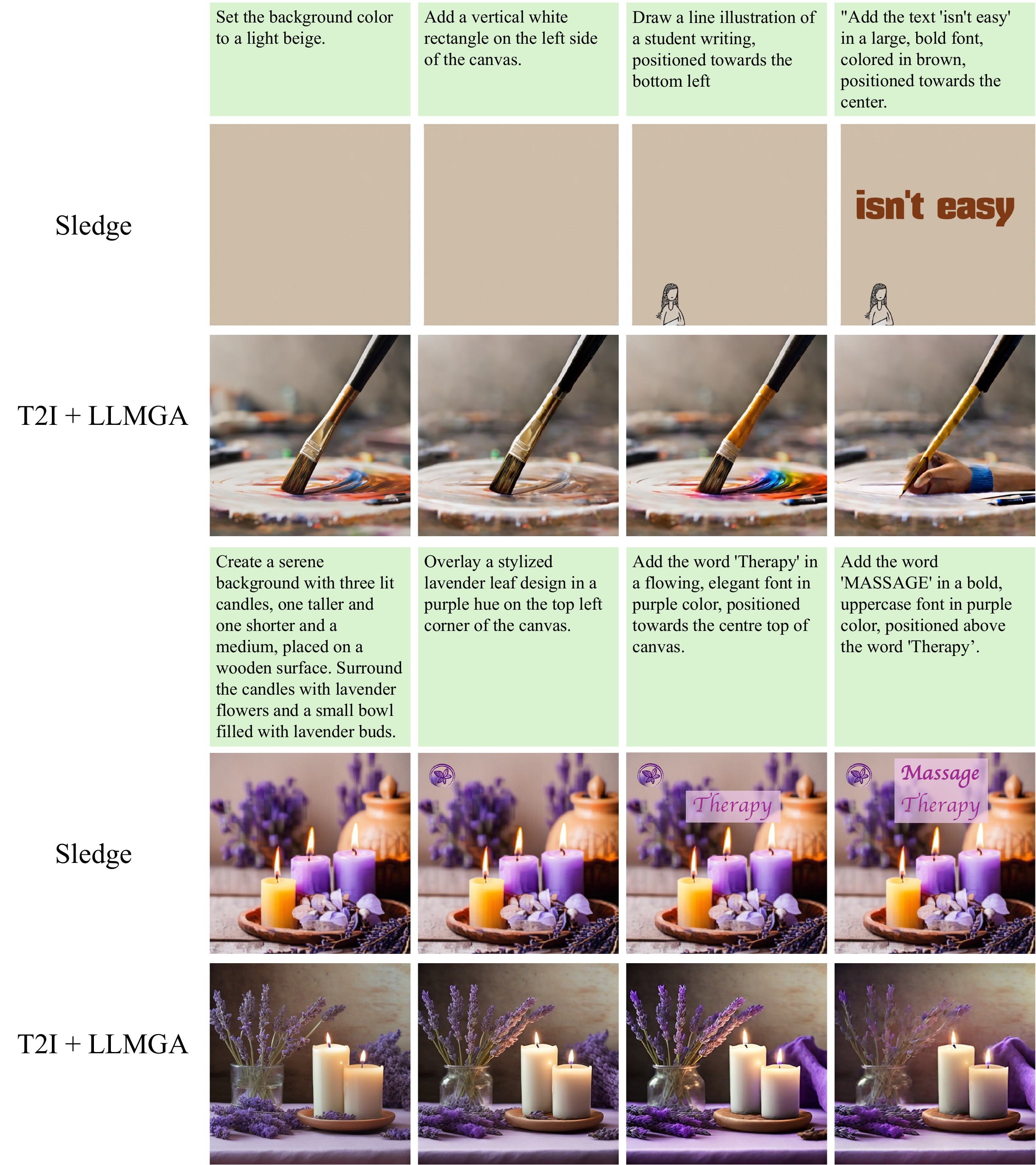}
  \caption{Qualitative comparison between our method and T2I+LLMGA for iterative design generation. In this figure, we showcase when T2I+LLMGA fails due to the discrepancy in the prompt refinement stage. (In the first example, a white rectangle of small width has been added on the left side) }
  \label{fig:qual3}
\end{figure*}

\begin{figure*}[t]
  \centering
  \includegraphics[width=0.9\textwidth]{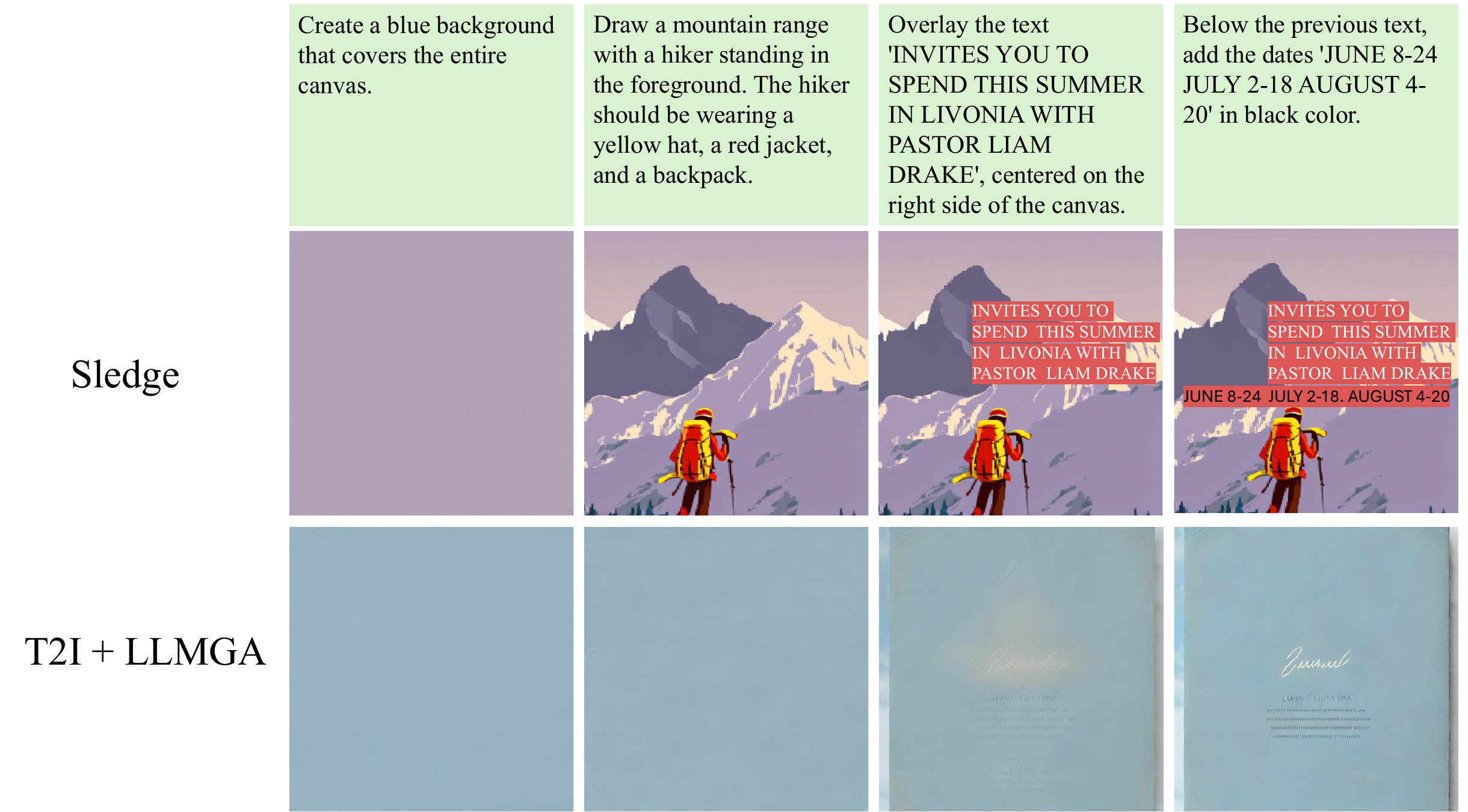}
  \caption{Qualitative comparison between our method and T2I+LLMGA for iterative design generation. This figure focuses on the editing part of the generation process, where our approach is able to follow human edits more closely.}
  \label{fig:qual1}
\end{figure*}

\begin{figure*}[t]
  \centering
  \includegraphics[width=0.9\textwidth]{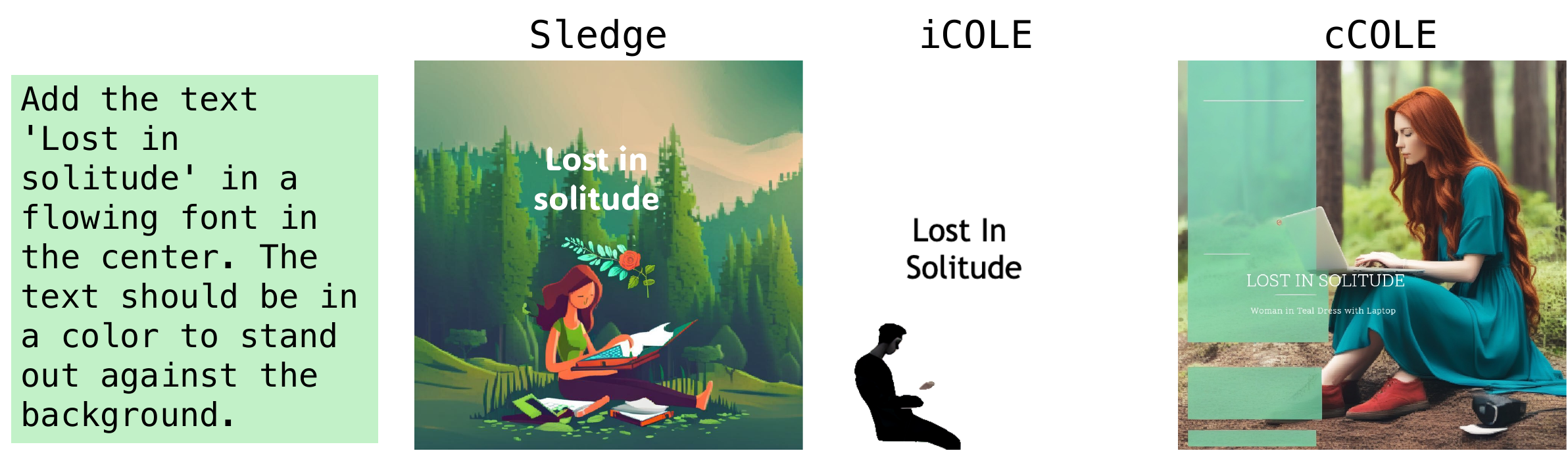}
  \caption{
 We present images where cCOLE fails to correctly predict the text. On the left, we display the original instruction, while on the right are the corresponding images generated by cCOLE. }
  \label{fig:ctext}
\end{figure*}

We demonstrate our method's iterative design generation capabilities through comparative examples with our strongest baseline T2I+LLMGA across three representative scenarios in~\cref{fig:qual1,fig:qual2,fig:qual3}.

Fig.~\ref{fig:qual2} reveals the baseline method's limitations, where its editing module fundamentally fails when challenged to introduce new elements to the existing canvas.
Furthermore, in~\cref{fig:qual3}, while the baseline generates visually aesthetic outputs, it demonstrates a disconnect between the generated output and the input instructions. We attribute this discrepancy to the LLMGA's refinement stage, wherein the Large Language Model (LLM) may misinterpret the user's original intent, leading to outputs that diverge substantially from the desired design. In~\cref{fig:qual1}, our proposed approach exhibits superior instruction comprehension by executing the user's edit requests.

\begin{figure*}[t]
  \centering
  \includegraphics[width=0.9\textwidth]{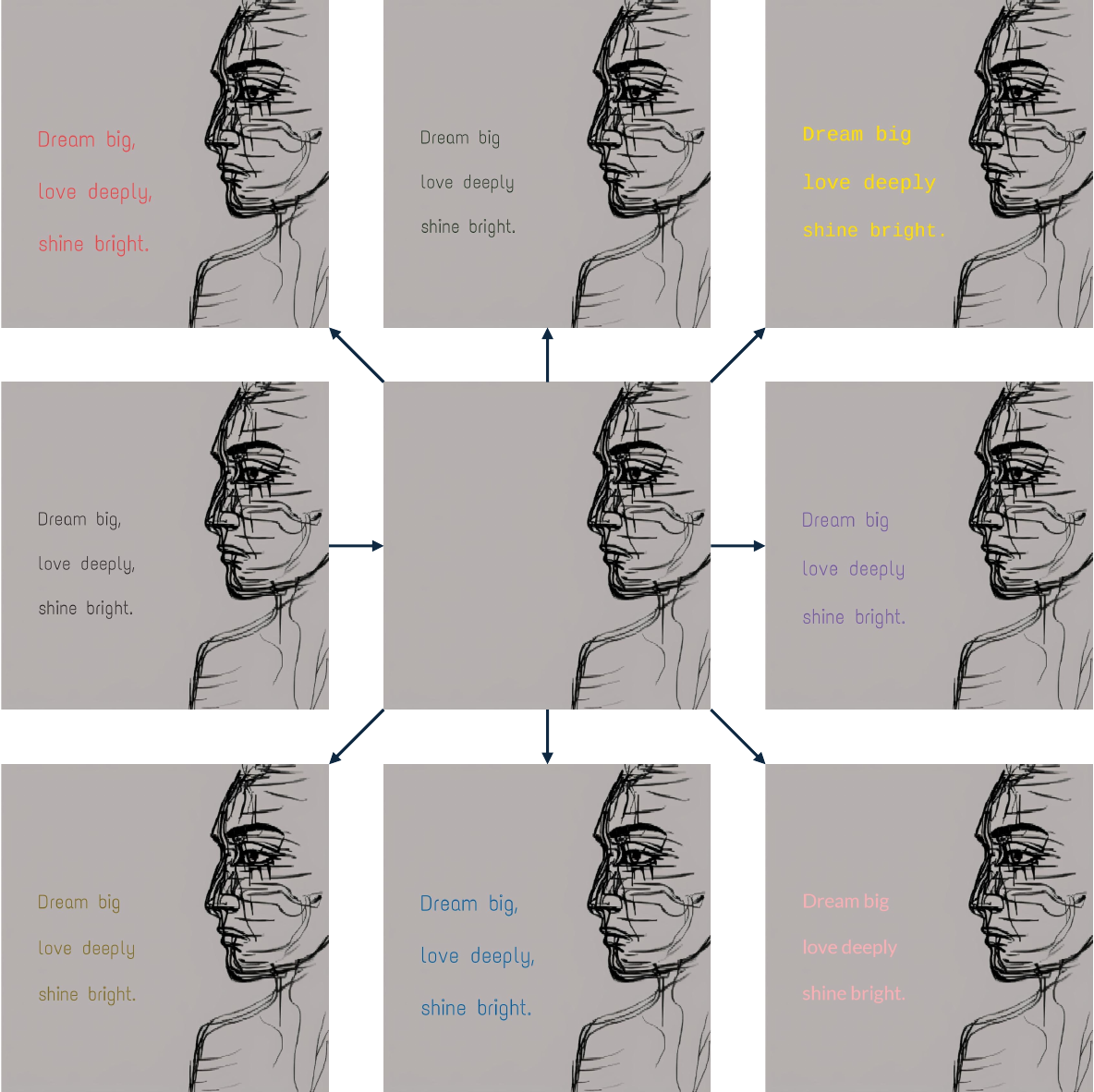}
  \caption{
  In these images, we show how our approach can make changes to the text attributes like color and font post-hoc. The layered output from our model facilitates this. This enhances the editability of the generated designs.
  }
  \label{fig:edit}
\end{figure*}

In~\cref{fig:edit}, we showcase how our approach can change the text contents in designs post-hoc. This is possible because, during the generation process, we generate layers of information and not a flat image. We showcase changing the font style and colors. This allows the designers to modify these attributes post-hoc to enhance the readability of the generations from the model.

\section{Circular evaluation and Prompts Used} \label{sec:prompt}

\subsection{Circular Evaluation}

When using MLLMs for comparative image evaluation, their responses can be influenced by ordering biases. This means that the position in which images are presented can affect the model’s judgment, leading to inconsistent or unfair comparisons.  To address this, we perform a circular evaluation approach, where each pair of images is evaluated twice, swapping their order in the second comparison. This ensures that any preference for one image over another is not simply a result of positional bias.  

\begin{itemize}
    \item Initial Comparison – The model is given two images (A and B) in a fixed order and asked to determine which is preferred based on a specific criterion (e.g., theme adherence, aesthetics). For edit compliance, we provide A1, A2, and B1, B2 along with the instruction.
    \item Order Flipping – The same two images are presented again, but their order is reversed: B and A (B1, B2, and A1, A2 for edit compliance). The model is asked the same question without knowing its previous answer.  
    \item Consistency Check - If the model provides the same preference in both cases (e.g., A is preferred over B in both orders), we record the result in favor of A. If the model changes its answer depending on the order (e.g., prefers A in the first case but B in the second), we label this as equal for both models.
\end{itemize}

With circular evaluation, we ensure that model decisions are not affected by the sequence in which images are presented.

\subsection{Promps Used}
In this section, we detail all the prompts used in our work. We first report the prompt used to create the \textit{IDeation Benchmark}, then we report all the different prompts used to evaluate the baseline methods and our method.

\subsection{IDeation Benchmark}
\textit{You are an expert designer tasked with creating step-by-step instructions for generating designs based on specific themes. Your instructions should start with a blank canvas and progressively build up to the final design rendition.} 
\textit{\noindent Here are a few examples of design instruction sequences: }

\vspace{.5cm}
\noindent\textit{Example 1: Modern Minimalist Living Room}
\begin{enumerate}
    \item \textit{Start with a blank white canvas representing an empty room.}
    \item \textit{Draw simple outlines for a large window on the far wall, letting in natural light.}
    \item  \textit{Add a sleek, low-profile gray sofa centered on the left wall.}
    \item \textit{Place a minimalist coffee table with a glass top in front of the sofa.}
    \item \textit{Include two accent chairs with clean lines on the right side of the room.}
    \item \textit{Hang a large, abstract painting with bold colors above the sofa.}
    \item \textit{Add a slim floor lamp with a metallic finish in the corner.}
    \item \textit{Place a simple area rug with a geometric pattern to tie the space together.}
\end{enumerate}

\vspace{.5cm}
\noindent\textit{Example 2: Birthday Party Invitation}
\begin{enumerate}
    \item \textit{Create a background with a crumpled foil texture, covering the entire canvas.}
    \item \textit{Place a white rectangle in the center of the canvas, covering a significant portion of the background.}
    \item  \textit{At the top of the white rectangle, add the word 'The' in a stylish, elegant font.}
    \item \textit{Below 'The', add the text 'GREAT BIRTHDAY' with 'GREAT' in red and 'BIRTHDAY' in black, using a bold and prominent font.}
    \item \textit{Directly below 'GREAT BIRTHDAY', add the word 'Party' in a cursive, elegant font.}
    \item \textit{Below 'Party', add the text 'Come to celebrate Harry's special day!' in a red, cursive font.}
    \item \textit{At the bottom of the white rectangle, add the text "Harry's House 851 E Kingsley Ave" in a simple, black font.}
\end{enumerate}

\vspace{.5cm}
\noindent \textit{Now, I will provide you with a theme. Based on this theme, generate a sequence of 8-10 step-by-step instructions for creating a design similar to the examples above. Each step should build upon the previous ones, starting from a blank canvas and ending with a complete design. The output should be in a Python dictionary format, meaning you should start with '\{' and end with '\}'. The theme of the design is <theme>.}

\subsection{Evaluation Prompts}
\textbf{Ideation Benchmark Instruction Filtering}: \textit{You are an expert in graphic design and instructional clarity. I have a set of step-by-step instructions that guide the creation of a design from a blank canvas to a completed image. These instructions cover diverse themes, such as workplace safety, holidays, and social media posts. Your task is to evaluate whether the instructions are coherent, logical, and meaningful. Specifically, check if the steps clearly build upon each other in a structured manner. The instructions are free from contradictions or missing steps that would hinder execution. The final design, as described by the instructions, makes sense and aligns with the intended theme. Respond with "Yes" if the instructions are coherent, meaningful, and can successfully lead to a well-structured design. Respond with "No", if the instructions are unclear, incomplete, illogical, or would lead to an incoherent design. Here are the instructions: <instructions>} \\
\textbf{Theme Matching Absolute}: \textit{Given the design rendition, how well does it match with the theme <theme>? Please rate the matching from 1 to 5, 1 means a poor match, 2 means fair match, 3 means a good match, 4 means very good match, and 5 means a perfect match; ensure that your response is only a single number and nothing else?} \\
\textbf{Aesthetic Quality Absolute}: \textit{From a designer point of view, given the image of a design rendition, please rate the design from 1 to 5, 1 meaning poor, 2 is fair, 3 is good, 4 is very good and 5 is perfect, reply only with a single number and nothing else?} \\
\textbf{Edit Compliance Absolute}: \textit{Does the second image follow the edit instruction provided to edit the first image? The instruction is <instruction>. Please rate how well it follows the edit instruction on a scale of 1 to 5. 1 meaning poor, 2 is fair, 3 is good, 4 is very good and 5 is perfect, reply only with a single number and nothing else?} \\
\textbf{Theme Matching Comparative}: \textit{Given the two design renditions, which one of them follows the theme <theme> better. reply only with Image1 or Image2 and nothing else?} \\
\textbf{Aesthetic Quality Comparative}: \textit{which image is better aesthetically to be used as a graphic design, reply only with Image1 or Image2 and nothing else?} \\
\textbf{Edit Compliance Comparative}: \textit{Given the edit instruction <instruction>. Does Image2 represent the edited image from Image1 according to the instruction, or does Image4 represent the edit better when edited from Image3? Choose which one of the images is edited better according to the instruction. Reply only with Image1 or Image3 and nothing else?}

\section{Dataset Examples} \label{sec:data_qual}
\begin{figure*}[t]
  \centering
  \includegraphics[width=0.9\textwidth]{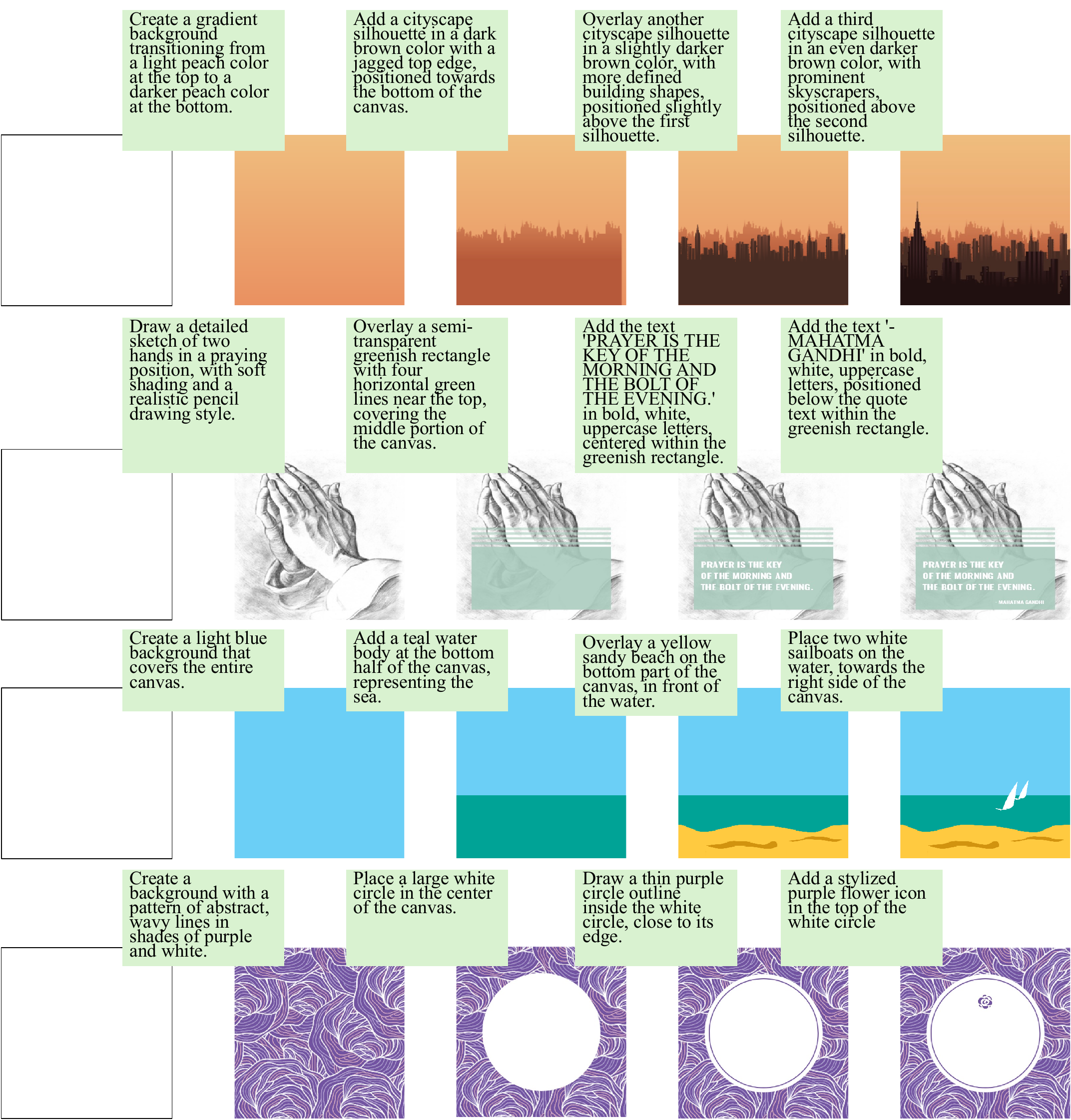}
  \caption{Examples from our proposed dataset showcasing step-by-step edit instructions. For clarity, we illustrate only five edit steps.}
  \label{fig:train}
\end{figure*}

In~\cref{fig:train}, we provide examples for the step-by-step edit instructions and the canvas state from our dataset. These examples show the complex edits involved in iterative design generation.

\section{Failure Cases} \label{sec:fail}
\begin{figure*}[t]
  \centering
  \includegraphics[width=0.9\textwidth]{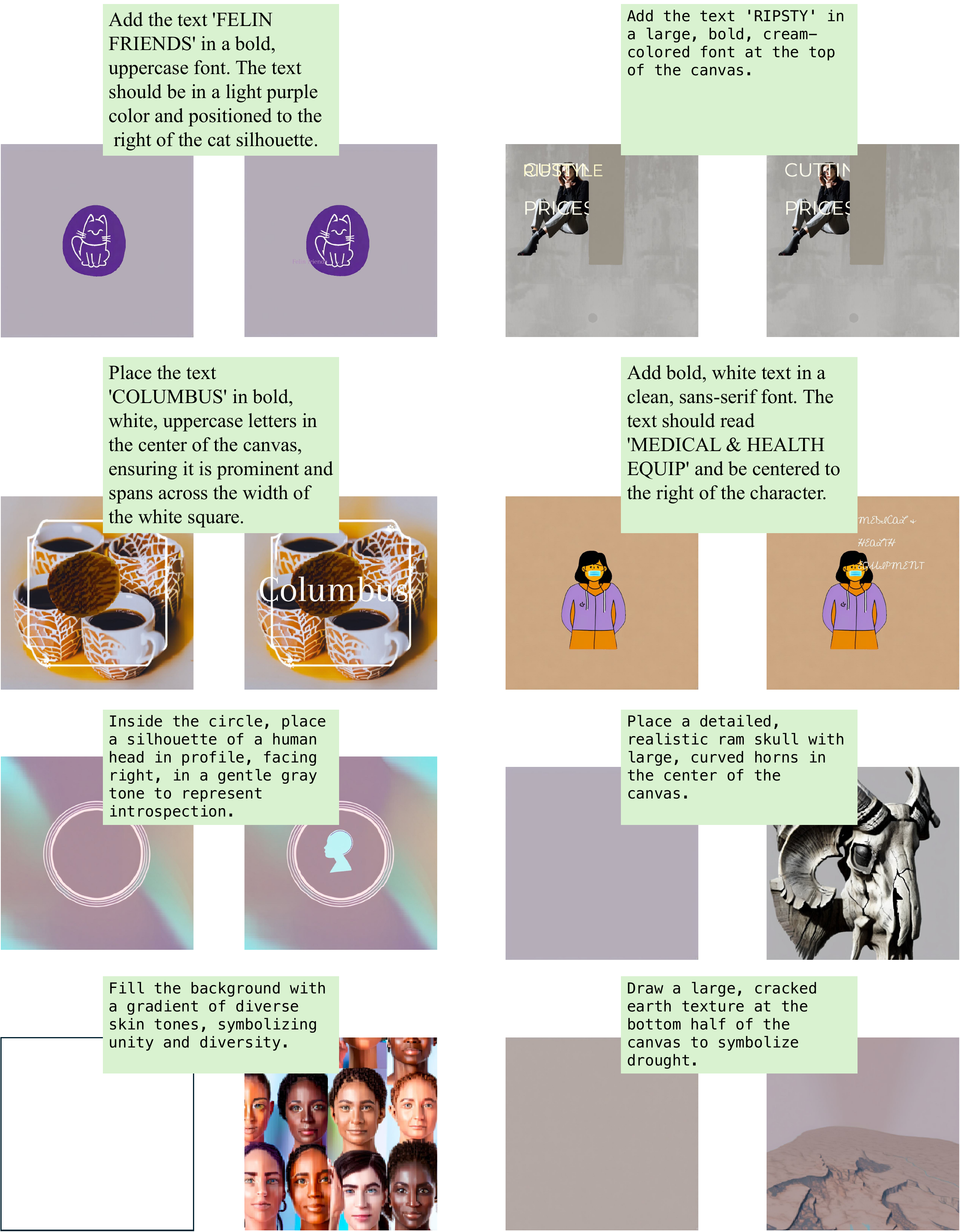}
  \caption{We highlight cases when our approach produces unsatisfactory results. In the top two rows, we show results with errors in predicting the textual location or size of the fonts, which affects the output. The bottom two rows showcase designs that, while following the instructions, fail to achieve the desired aesthetic quality.}
  \label{fig:fail}
\end{figure*}

Our method outperforms baseline approaches in most scenarios. However, in this section, we focus on analyzing cases where our method fails to produce the intended outputs. 1) Misplacement of $\vU_{t+1}$: In certain instances, the MLLM fails to accurately predict the location of $\vU_{t+1}$, leading to suboptimal placement of text. This issue is illustrated in the first two rows of~\cref{fig:fail}. 2) Lack of Aesthetic Appeal: In the bottom two rows of~\cref{fig:fail}, we observe cases where, despite adhering to the given instructions, the generated outputs lack visual appeal and fail to deliver aesthetically pleasing renditions.

\newpage

\end{document}